\documentclass[journal]{IEEEtran}
\usepackage{algorithmic}
\usepackage{graphicx}
\usepackage{subfigure}
\usepackage{url}
\usepackage{cite}
\usepackage{verbatim} 

\usepackage{amsmath}
\usepackage{amssymb}

\usepackage{color}
\usepackage{xcolor}
\usepackage{threeparttable}
\usepackage{makecell}
\usepackage{nicefrac}
\usepackage{hyperref}

\usepackage{pifont}
\newcommand{\xmark}{\ding{53}}%

\newcommand{\blue}[1]{\textcolor{black}{#1}}

\begin{document}

\title{\blue{Outracing Human Racers with Model-based Planning and Control for Time-trial Racing}}

\author{Ce Hao$^1$,
        Chen Tang$^1$,
        Eric Bergkvist$^2$,
        Catherine Weaver$^1$,
        Liting Sun$^1$,
        Wei Zhan$^1$,
        Masayoshi Tomizuka$^1$
        
\thanks{$^1$ C. Hao, C. Tang, C. Weaver, L. Sun, W. Zhan and M. Tomizuka are with the Department of Mechanical Engineering, University of California Berkeley, CA, USA \{cehao, chen\_tang, catherine22, litingsun, wzhan, tomizuka\}@berkeley.edu.}
\thanks{$^2$ E. Bergkvist is with École polytechnique fédérale de Lausanne (EPFL), Switzerland (eric\_bergkvist@yahoo.com). This work was done during his visit at University of California Berkeley.}
\thanks{This work was supported by Sony R\&D Center Tokyo and Sony AI.}

}


\maketitle
\begin{abstract}
Autonomous racing has become a popular sub-topic of autonomous driving in recent years. The goal of autonomous racing research is to develop software to control the vehicle at its limit of handling and achieve human-level racing performance. In this work, we investigate how to approach human expert-level racing performance with model-based planning and control methods using the high-fidelity racing simulator Gran Turismo Sport (GTS). GTS enables a unique opportunity for autonomous racing research, as many recordings of racing from highly skilled human players can serve as expert demonstrations. By comparing the performance of autonomous racing software with human experts, we better understand the performance gap of existing software and explore new methodologies in a principled manner. In particular, we focus on the commonly adopted model-based racing framework, consisting of an offline trajectory planner and an online Model Predictive Control-based (MPC) tracking controller. We thoroughly investigate the design challenges from three perspectives, namely vehicle model, planning algorithm, and controller design, and propose novel solutions to improve the baseline approach toward human expert-level performance. We showed that the proposed control framework can achieve top $0.95\%$ lap time among human-expert players in GTS. Furthermore, we conducted comprehensive ablation studies to validate the necessity of the proposed modules, and pointed out potential future directions to reach human-best performance. 
\end{abstract}

\begin{IEEEkeywords}
Autonomous racing, trajectory planning, model predictive control
\end{IEEEkeywords}

\section{Introduction} \label{Sec:Introduction}
Inspired by racing competitions such as Formula One, Rallying, and IndyCar, \emph{autonomous racing} has emerged recently as an important sub-field of autonomous driving research~\cite{betz2022autonomous, paden2016survey, chen2022milestones}. One fundamental problem of interest is time-trial competition, where each race car drives solo around a track and competes to achieve the fastest lap time. Time-trial autonomous racing requires the developed control software to operate the race car at extremely high speeds and the dynamic limits of handling. While expert human racers are proficient in pushing the race car to its limits of handling, this task is extremely challenging for current autonomous driving software, as it is designed primarily for regular driving conditions \blue{(e.g. urban and highway)}. Therefore, one important research question is how to design autonomous racing systems that can achieve the same level of performance or even outperform the best human experts. By solving this problem, we could improve current autonomous driving systems by increasing robustness and safety under extreme conditions. 

We aim to compare the performance of autonomous race cars with skilled human racers using the same environment. Control algorithms developed in existing works are often evaluated on full size\cite{kapania2016trajectory}, miniature\cite{hewing2018cautious}, and simulated\cite{wymann2000torcs} race cars. The experimental results of such prior works provide informative insights into the performance of the control algorithms; however, it is difficult to estimate the performance gap between a given algorithm and human experts on many of these testing platforms, as collecting true ``expert'' demonstrations using a comparable environment to the control platform may be extremely difficult. In this work, we develop and evaluate our algorithm on a racing simulator that enables principled comparison between the performance of autonomous race cars and human experts\textemdash Gran Turismo Sport (GTS)\cite{GTS_web}. GTS simulates vehicles with high-fidelity vehicle dynamics models. More notably, GTS is played by millions of players\footnote{See \url{http://www.kudosprime.com}}, many of whom are extremely skilled, so recordings of expert demonstrations in the GTS environment are readily available. This allows us to directly compare the closed-loop trajectories of autonomous race cars with the recordings of human experts on the \emph{same track} and with the \emph{same car}. 

\begin{figure}[t]
    \centering
    \includegraphics[width=\columnwidth]{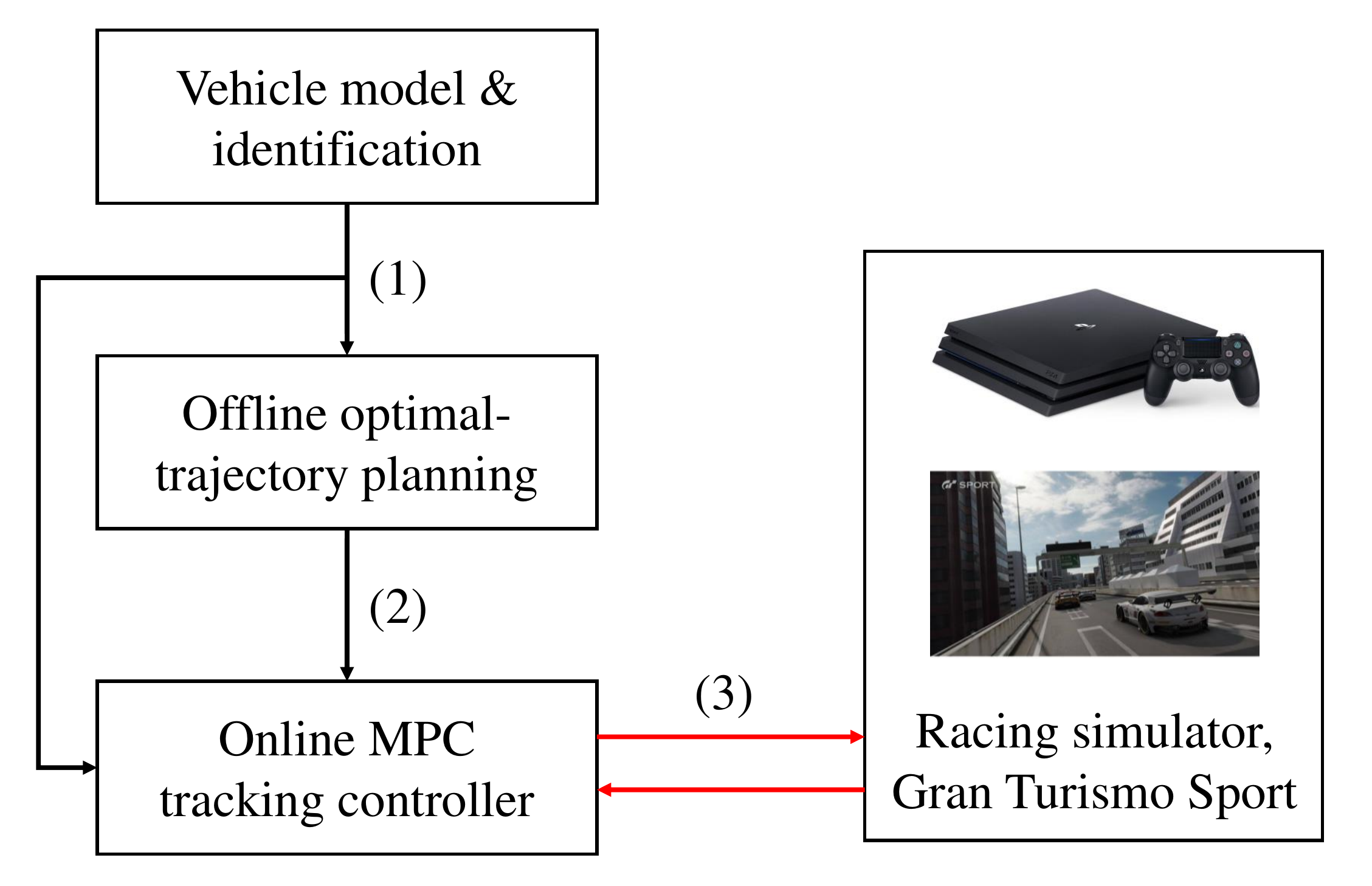}
    \caption{The overall framework of the racing control system. \textbf{(1)} The vehicle model provides the dynamic model constraint in offline planning and online control. \textbf{(2)} The online controller tracks the planned time-optimal trajectory to achieve human expert-level performance. \textbf{(3)} Experiments are conducted in the high-fidelity GTS simulator. 
    }
    \label{fig:overall_framework}
\end{figure}

In particular, we are interested in studying how to approach human-level performance with \emph{model-based control methods}. While prior works~\cite{wurman2022outracing, fuchs2021super} have shown that model-free reinforcement learning (RL) can achieve super-human performance in the GTS environment, these RL agents rely on iterative sampling over many trials\textemdash the champion-level RL racing agent, GT Sophy~\cite{wurman2022outracing}, was trained on 10 Play Stations for \emph{120 billion} training steps which took a whole week. Notably, the opaque nature of deep neural networks prevents principled analysis of the learned policies. In contrast, in this work, we aim to compose an interpretable model-based control framework to study this problem in a structured manner. We are interested in identifying the particular elements of the autonomous racing problem that are crucial to approach human-level lap time. 

We consider a commonly adopted model-based racing control framework. As shown in Fig.~\ref{fig:overall_framework}, it consists of three modules: 1) vehicle models identified from driving data collected in GTS; 2) a trajectory planning module that finds the time-optimal trajectory \emph{offline}; and 3) a model-based controller that tracks the time-optimal reference \emph{online}. We thoroughly investigate the design challenge of each module and propose novel solutions to improve current state-of-the-art approaches toward human-level performance. With our proposed control framework, we achieved the \emph{top $1\%$} lap time among human players. We conducted comprehensive ablation studies to validate the necessity of each design decision. Also, we compared the results with the best human recordings to pinpoint the remaining challenges for future research. Here we give a brief summary of our improvements to each module and defer the detailed discussion to the remaining sections. 

{\bf Vehicle Models.} An accurate vehicle model is a prerequisite for good model-based planning and control performance. However, a complex model is infeasible for practical control design. The bicycle model~\cite{rajamani2011vehicle, paden2016survey, subosits2021impacts} can approximate the chassis movement well and is widely adopted in vehicle control. However, the nominal bicycle model is insufficient to describe vehicle dynamics under extreme conditions. In our experiments, we found that two dynamic effects neglected in prior works~\cite{rajamani2011vehicle, rosolia2017autonomous, kong2015kinematic}, aerodynamic forces and weight transfer, are crucial for accurate prediction of vehicle motion in both planning and control to achieve minimum lap time. 

{\bf Offline Planning.} To achieve minimum lap time, the offline planning module needs to provide an optimal reference trajectory for tracking. But the lap time is a highly nonlinear objective to optimize. As a result, it is challenging to solve the time-optimal problem directly~\cite{metz1989near, heilmeier2019minimum}. In the literature, people simplified the time-optimal objective and derived the quadratic curvature-optimal objective. While the resulting optimization problem is easy to solve, the solution may be neither optimal nor feasible for the race vehicle to track, which prevents autonomous race cars from reaching human expert-level lap time. To this end, we argued that lap time should still be the primary objective. To ensure the time-optimal problem can be solved stably, we proposed to warm-start it with a curvature-optimal trajectory.

{\bf Online Tracking Controller.} In this work, we used Model Predictive Control (MPC) to design the low-level tracking controller. MPC has been widely adopted for vehicle control in both urban and racing scenarios~\cite{abbas2011non, carvalho2013predictive, chen2019autonomous, rosolia2017autonomous, ming2016mpc, rafaila2015nonlinear, mata2019robust}. In most applications, the control problem is decoupled into longitudinal and lateral control for simplicity~\cite{amati2019lateral}. However, the longitudinal and lateral motions are tightly coupled for high-speed race cars at sharp corners. Consequently, the decoupled controllers lead to large tracking errors and sub-optimal performance. To this end, we designed an MPC tracking controller that jointly optimizes the control actions based on the longitudinal and lateral dynamics of the vehicle model. We found that this coupled formulation of the MPC controller significantly reduces tracking errors and lap time. It played an important role in achieving human expert-level lap time.

\blue{\textbf{Contribution} of this paper is threefold: 1) we implement a  model-based racing control system on the high-fidelity racing simulation Gran Turismo Sport and achieve expert-level performance; 2) we conducted comprehensive experiments and ablation studies to analyze the function of each module in the control system and the racing performance on the corners of various shapes; 3) we propose a two-stage time-optimal trajectory planning method warm-started by the curvature-optimal trajectory.} The rest of the paper is organized as follows. In Sec.~\ref{Sec:RelatedWork}, we review the related literature on vehicle models, planning, and control algorithms. In Sec.~\ref{Sec:VehicleModel}, \ref{Sec:TrajPlan} and \ref{Sec:Controller}, we introduce the key improvements we made in the three modules respectively. In Sec.~\ref{Sec:ExpResult}, we report a comprehensive set of experiments and analyses to validate the effectiveness of the proposed method. 

\begin{figure*}[t]
    \centering
    \includegraphics[width = 0.8\textwidth]{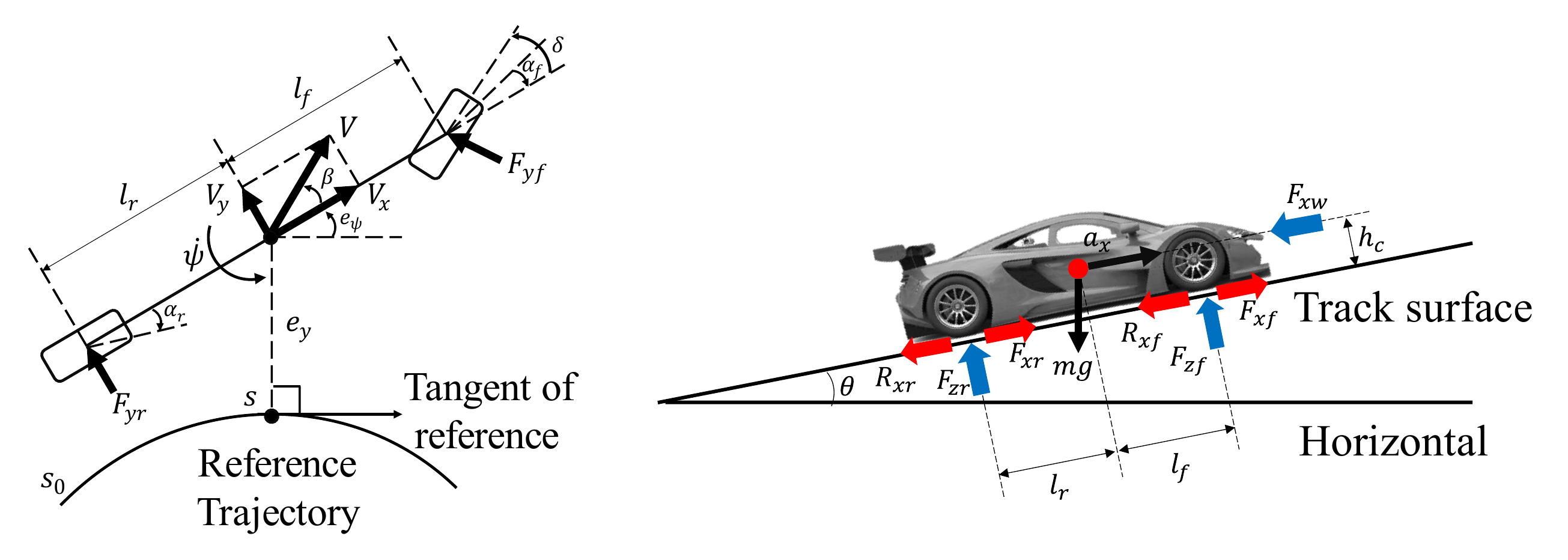} 
    \caption{Diagram of race vehicle model. \textbf{Left}: Overview and relative position to the reference trajectory; \textbf{Right}: Lateral profile with track surface pitch angle.}
    \label{fig:vehicle model}
\end{figure*}

\section{Related Works} 
\label{Sec:RelatedWork}
The recent advances in autonomous racing have been summarized comprehensively in Betz \textit{et al.}~\cite{betz2022autonomous}. In this section, we only give a concise review of some related works on racing dynamics modeling, model-based trajectory planning, and control methods for time-trial racing games. 

\subsection{Vehicle Models}
The vehicle motion is described by the combination of the kinematic and the dynamic model~\cite{rajamani2011vehicle, rosolia2017autonomous, kong2015kinematic}. For racing in closed tracks, a common practice is to define the kinematic model in the Frenet-Serret coordinate~\cite{rosolia2019learning} regarding the track centerline so that we can conveniently obtain a differentiable model of the positional and heading errors on curvy tracks. For the dynamics model, the bicycle model~\cite{rajamani2011vehicle} is the most widely adopted model used in vehicle control. 

In this work, we formulate the vehicle model based on the widely used bicycle model~\cite{rajamani2011vehicle}. In particular, we would like to emphasize the necessity of two dynamic effects\textemdash aerodynamic force and weight transfer. While they are normally neglected in urban autonomous driving and prior works in autonomous racing~\cite{rajamani2011vehicle, rosolia2017autonomous, kong2015kinematic}, they are important for achieving minimum lap time and reaching human expert-level performance. The longitudinal dynamics is mainly governed by the pedal command through the powertrain dynamics~\cite{rajamani2011vehicle}, whereas the lateral dynamics is governed by the lateral tire forces. A popular option for tire force modeling in vehicle control is Pacejka's Magic formula\cite{pacejka2005tire, milliken1995race}. Aerodynamic force and weight transfer are two effects that are well described in the literature of vehicle dynamics~\cite{milliken1995race, rucco2015efficient}. However, they are normally neglected in prior works in both urban autonomous driving and autonomous racing~\cite{rajamani2011vehicle, rosolia2017autonomous, kong2015kinematic}. 

\subsection{Trajectory Planning Algorithms} 
\blue{Linger \textit{et al.} employed model predictive contouring control to maximize the traveling distance by optimizing the local reference trajectory~\cite{liniger2015optimization}. Ugo \textit{et al.} proposed learning MPC to iteratively plan local optimal path with the least steps~\cite{rosolia2017autonomous}. However, optimizing the trajectory locally is not sufficient for minimizing the overall lap time.} In our proposed framework, the planning module finds the optimal racing line of the whole track in an \emph{offline} manner. Two categories of approaches are commonly adopted in the literature. Time-optimal planning methods~\cite{velenis2005minimum, christ2021time} directly use the lap time as the objective to optimize. Vázquez \textit{et al.}~\cite{vazquez2020optimization} wrote the lap time as a non-convex function of vehicle states in the Frenet coordinate. The non-convex objective function makes the optimization problem difficult to solve. Metz \textit{et al.} \cite{metz1989near} solved the time-optimal optimization via quasi-linearization. Christ \textit{et al.}~\cite{christ2021time} approximated the time-optimal objective as low-order polynomials using Gauss-Legendre collection to reduce the local minima problem in the nonlinear objective. \blue{Nevertheless, the model linearization introduces errors and makes the planned trajectory inaccurate.}

Another line of methods chooses curvature as the objective function. The motivation is to minimize the curvature of the racing line in corners, allowing the race cars to pass the corners at high speed. Heilmeier \textit{et al.} ~\cite{heilmeier2019minimum} proposed an approximated geometry expression of path curvature, which was used as the objective function of a nonlinear programming problem. Kapania \textit{et al.}~\cite{kapania2016sequential} proposed a two-step iterative algorithm to decompose the curvature-optimal problem into two convex optimization problems, which were solved in turn to optimize the path curvature and the speed profile, respectively. While the curvature-optimal problem is easier to solve, the solution is suboptimal in terms of lap time. It cannot satisfy our objective of reaching top human-level performance.

\subsection{Model Predictive Control for Racing}
There is a rich literature on using model predictive control to design vehicle controllers for autonomous driving~\cite{abbas2011non, carvalho2013predictive, chen2019autonomous, rosolia2017autonomous, ming2016mpc, rafaila2015nonlinear}. A common practice is to decouple the longitudinal and lateral controls for simplicity~\cite{kong2015kinematic}. However, as we will show in our experimental study, such a decoupled MPC controller leads to large tracking errors and poor overall performance in racing scenarios. For example, researchers \cite{rosolia2019learning, kabzan2019learning} adopted MPC controllers based on the coupled longitudinal and lateral dynamics model. Our contribution in this aspect lies in providing a practical coupled MPC controller design and showing its crucial role in reaching top human-level performance through a detailed experimental study and analysis. 

\section{race vehicle Model}
\label{Sec:VehicleModel}
In this section, we present the vehicle model used in our racing control framework. As mentioned before, we need an accurate vehicle model to ensure good model-based planning and control performance. However, a high-fidelity but complex model, such as the one introduced in~\cite{milliken1995race}, is prohibitive for accurate identification and efficient real-time control. Therefore, a key design decision is to make an appropriate trade-off between modeling accuracy and complexity. In this work, we formulate the vehicle model based on the widely used bicycle model~\cite{rajamani2011vehicle}. In particular, we would like to emphasize the necessity of two dynamic effects\textemdash aerodynamic force and weight transfer. While they are normally neglected in urban autonomous driving and prior works in autonomous racing~\cite{rajamani2011vehicle, rosolia2017autonomous, kong2015kinematic}, they are important for achieving minimum lap time and reaching human expert-level performance. 

The overall vehicle model is depicted in Fig.~\ref{fig:vehicle model}. We consider a front-wheel-steering and rear-wheel-drive car. We model its motion regarding a reference trajectory in the Frenet coordinate. \blue{
We define the centerline of the track and the planned time-optimal trajectory as the reference lines in the Frenet Coordinate for planning and control. In offline planning, we employ Frenet-Serret coordinates with respect to the curve of the track centerline. In closed-loop control, we employ Frenet-Serret coordinates with respect to the curve of the offline-planned trajectory.} 
The states and control inputs are defined as:
\begin{equation}\label{Eqn:states_and_controls}
    \xi = [V_x, V_y, \dot{\psi}, e_\psi, e_y, s]^T, \quad u = [\delta, a_x]^T,
\end{equation}
where $V_x$ and $V_y$ denote the longitudinal and lateral velocities, $\dot{\psi}$ denotes the yaw rate, $e_{\psi}$ and $e_y$ denote the relative yaw angle and distance in the Frenet coordinate, $s$ denotes the traveling distance along the reference trajectory from an initial point $s_0$, $\delta$ denotes the steering angle, and $a_x$ denotes the longitudinal acceleration generated by tire forces~\cite{kong2015kinematic}. 

\subsection{Kinematic Model in Frenet Coordinate}
We describe the vehicle kinematic model in the Frenet-Serret coordinate system, which describes motion with respect to a curve. The state variables $e_\psi, e_y, s$ are defined regarding the point on the curve that is the closest to the center of gravity (CoG). Following~\cite{rosolia2019learning}, we model the vehicle kinematics as follows:
\begin{align}
\dot{e}_{\psi}&=\dot{\psi}-\frac{V_{x} \cos \left(e_{\psi}\right)-V_{y} \sin \left(e_{\psi}\right)}{1-\kappa(s) e_{y}} \kappa(s), \label{Eqn:epsi} \\
\dot{e}_{y}&=V_{x} \sin \left(e_{\psi}\right)+V_{y} \cos \left(e_{\psi}\right),\label{Eqn:ey}\\
\dot{s}&=\frac{V_{x} \cos \left(e_{\psi}\right)-V_{y} \sin \left(e_{\psi}\right)}{1-\kappa(s) e_{y}}, \label{Eqn:s}
\end{align}
where $\kappa$ denotes the curvature of the closest point on the curve. 

\subsection{Vehicle Chassis Dynamic Model}
Here we present the chassis dynamic model describing the vehicle dynamics along the longitudinal and lateral directions. The differential equations are given by:
\begin{align}
\dot{V}_{x}= & a_x - \frac{1}{m}\left(F_{y f} \sin (\delta)+R_{x}+F_{x w}\right)  \nonumber
 \\  & -g \sin (\theta)+ \dot{\psi} V_{y}, \label{Eqn:long_model}\\ 
\dot{V}_{y}=&\frac{1}{m}\left(F_{y f} \cos (\delta) +F_{y r}\right)-\dot{\psi} V_{x}, \label{Eqn:lat_model1} \\
\ddot{\psi}=&\frac{1}{I_{z z}}\left(l_{f} F_{y f} \cos (\delta)-l_{r} F_{y r}\right), \label{Eqn:lat_model2}
\end{align}
where $m$ and $I_{zz}$ denote the mass and moment of inertia, respectively, and  $l_f$ and $l_r$ represent the distance from the CoG to the front and rear axles, respectively. We describe the longitudinal dynamics, i.e., Eq.~\eqref{Eqn:long_model}, and the lateral dynamics, i.e., Eq.~\eqref{Eqn:lat_model1}-\eqref{Eqn:lat_model2}, respectively, in the following paragraphs. 

{\bf Longitudinal dynamics.} The longitudinal dynamics of the vehicle arise from a force-mass balance in the longitudinal direction, i.e., Eq.~\eqref{Eqn:long_model}, and are primarily influenced by the longitudinal acceleration from tire forces $a_x$. In addition, longitudinal acceleration is also affected by several other forces. First, since the vehicle is front-wheel-steering, the lateral tire force of the front tires, $F_{yf}$, has a component along the longitudinal direction when the vehicle is steering. Moreover, acceleration due to gravity, $g$, also contributes to the longitudinal acceleration when the track has a non-zero pitch angle, $\theta$. The rolling resistance, $R_x$, exerted on both tires also damps the longitudinal acceleration\cite{rajamani2011vehicle}. More importantly, the \emph{wind drag force}, $F_{xw}$, generated by the vehicle slipstream significantly affects the longitudinal dynamics of race cars. The magnitude of the wind drag force is proportional to the square of longitudinal velocity~\cite{rajamani2011vehicle}: 
\begin{align}
&F_{x w}=C_{x w} V_{x}^{2}, \label{Eqn:Fxw} 
\end{align}
where $C_{xw}$ is a coefficient. Therefore, the wind drag force is significantly larger for race cars that operate at high speed, in contrast to vehicles under normal driving conditions. 

{\bf Lateral dynamics.}
The lateral dynamics result from a force-mass balance along the lateral direction, i.e., Eq.~\eqref{Eqn:lat_model1} and a moment-inertia balance around the vertical axis, i.e., Eq.~\eqref{Eqn:lat_model2}, and are primarily influenced by the lateral tire forces on the front and rear tires, $F_{yf}$ and $F_{yr}$ respectively. In this work, we adopt the simplified Pacejka’s Magic Formula~\cite{pacejka2005tire} to model the lateral tire forces as 
\begin{align} 
    F_{yf} &= \mu F_{zf} D_f \cdot \sin\left(C_f \cdot \arctan(B_f \cdot \alpha_f)\right), \label{Eqn:tire1}\\
    F_{yr} &= \mu F_{zr} D_r \cdot \sin\left(C_r \cdot \arctan(B_r \cdot \alpha_r)\right), \label{Eqn:tire2},
\end{align}
where $B$, $C$, and $D$ are model coefficients. The coefficient of friction, $\mu$, is determined by the road surface and tire. The wheel loads of the front and rear tire are $F_{zf}$ and $F_{zr}$, respectively. Pacejka’s Magic Formula relates the lateral tire forces to the slip ratios of the front and rear tires, $\alpha_f$ and $\alpha_r$,  defined as \cite{rajamani2011vehicle}
\begin{equation}
    \alpha_{f}=\delta-\frac{V_{y}+l_{f} \dot{\psi}}{V_{x}},\quad \alpha_{r}=-\frac{V_{y}-l_{r} \dot{\psi}}{V_{x}}.
\end{equation}

Accurate modeling of the vertical wheel loads is important, as wheel load determines the maximum lateral traction provided by each tire. Wheel loads are approximately constant for vehicles operating at low speed; however, in racing, \emph{the transfer of wheel loads} between the front and rear tires can be critical for race cars. Longitudinal weight transfer occurs when large longitudinal acceleration leads to pitch motion. The wheel loads $F_{zf}$ and $F_{zr}$ are affected by the longitudinal acceleration as follows: 
\begin{align}
\Delta W &= \frac{m \dot{V}_{x} h_{c}}{l_{f}+l_{r}}, \\
F_{z f} &=\frac{m (g\cos (\theta)+a_z) l_{r}}{l_{f}+l_{r}} - \Delta W, \label{Eqn:Fzf} \\
F_{z r} &=\frac{m (g\cos (\theta)+a_z) l_{f} }{l_{f}+l_{r}} + \Delta W, \label{Eqn:Fzr} 
\end{align}
where $\Delta W$ is the longitudinal wheel load transfer and $h_c$ denotes the height of CoG. The term $m \dot{V}_{x} h_{c}$, which accounts for the effect of weight transfer, is significant when the race car undergoes large acceleration or deceleration, often occurring when the race car enters and leaves sharp corners. As a result, longitudinal weight transfer can contribute to under-steering or over-steering at those sharp corners. Therefore, while previous works may have neglected the effect of weight transfer ~\cite{rajamani2011vehicle, rosolia2017autonomous, kong2015kinematic}, \blue{precisely modeling cars' turning ability crucially improves the racing performance.}

\blue{
In addition, wheel load may also transfer laterally between two front and rear wheels when the vehicle turns heavily. In the double-track model~\cite{jin2019advanced, christ2021time, veneri2020free}, the lateral wheel load transfer is modeled as,
\begin{equation} \label{Eqn: lateral weight transfer}
\begin{aligned}
    &\Delta W_{F} = \frac{m\dot{V}_y h_c}{l_t} \frac{l_r}{l_f+l_r},  \quad
    \Delta W_{R} = \frac{m\dot{V}_y h_c}{l_t} \frac{l_f}{l_f+l_r}, \\
    &F_{zFL} = \frac{F_{zf} - \Delta W_{F}}{2}, \quad F_{zFR} = \frac{F_{zf} + \Delta W_{F}}{2}, \\
    &F_{zRL} = \frac{F_{zf} - \Delta W_{R}}{2}, \quad F_{zRR} = \frac{F_{zf} + \Delta W_{R}}{2},
\end{aligned}
\end{equation}
where $l_t$ denotes track width; $\Delta W_F$ and $\Delta W_R$ are lateral wheel load transfer; $F_{zFL}, F_{zFR}, F_{zRL}, F_{zRR}$ denote the modified wheel load on Front/Rear and Left/Right wheels. However, the lateral load transfer has less impact on the lateral velocity and yaw rate according to the lateral dynamics. More details are discussed in Appendix~\ref{Appendix: 4wheel model}.}

\section{Trajectory Planning}
\label{Sec:TrajPlan}
The goal of trajectory planning is to find a trajectory that minimizes the vehicle's lap time around the track. However, direct minimization of lap time subject to the constraints of the nonlinear vehicle dynamics model is a non-convex optimization problem. Current solvers for non-convex optimization are extremely sensitive to initial guesses. Depending on the initial guess, the optimization process may end up failing to find a feasible trajectory or get stuck in bad local minima~\cite{bitar2019warm, zhang2018autonomous}. 

\begin{figure}[t]
    \centering
    \includegraphics[width=\columnwidth]{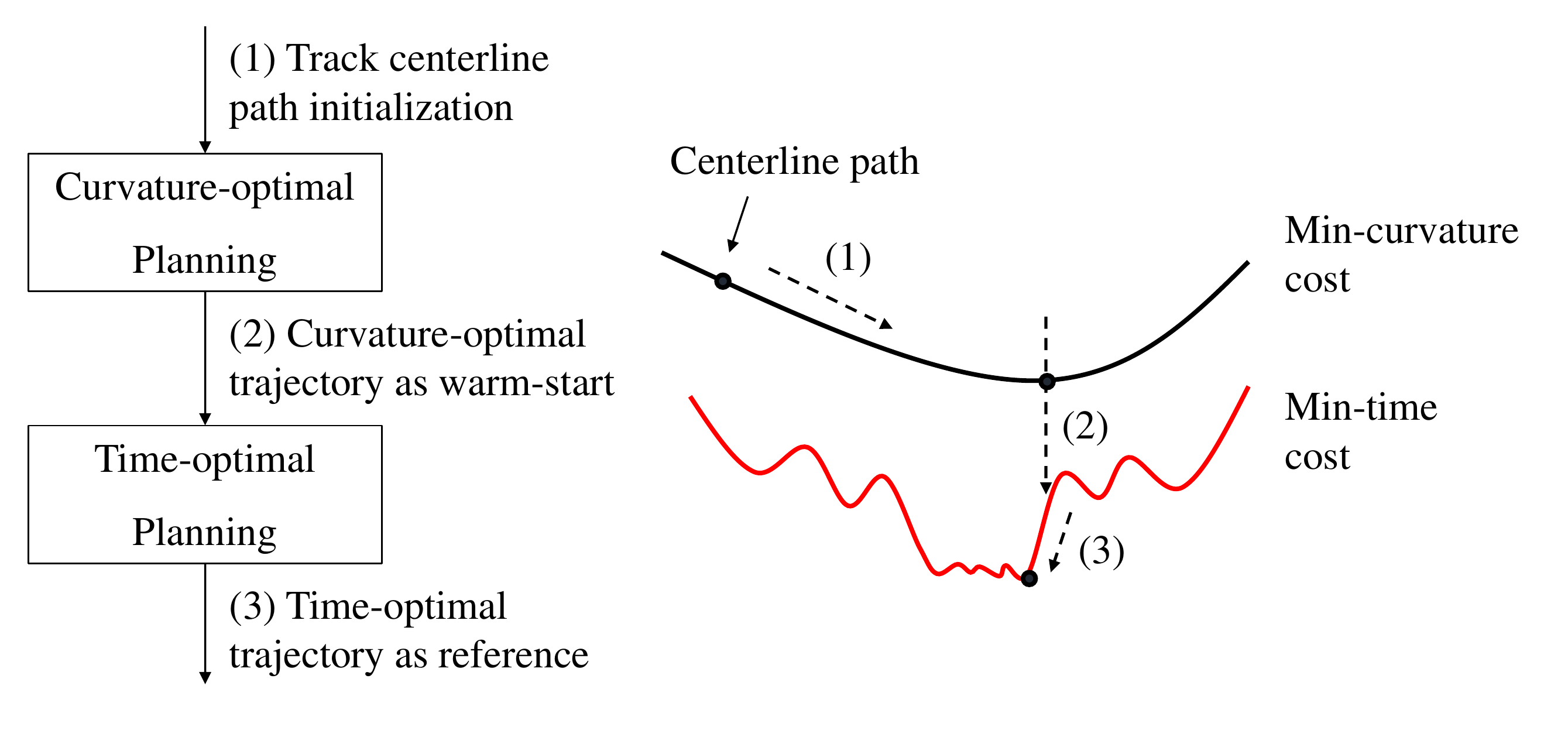}
    \caption{\textbf{Left}: The two-stage time-optimal planning framework. 
    \textbf{Right}: The illustration of cost minimization. 
    \textbf{(1)} Initialize the iterative curvature-optimal planning with the track centerline; \textbf{(2)} Solve the convex problem to the optimal solution, feed in the nonlinear solver with the curvature-optimal trajectory; 
    \textbf{(3)} The nonlinear solver solves the time-optimal planning problem after warm-started with the curvature-optimal trajectory.}
    \label{Fig:minkap}
\end{figure}

To tackle this problem, we propose a \textit{two-stage trajectory planning algorithm} detailed in Fig.~\ref{Fig:minkap}. In the first stage, instead of directly solving the non-convex time-optimal planning problem, we sought to find the minimum-curvature path subject to linearized vehicle dynamics first. Since the resulting curvature-optimal problem is convex, it can be reliably solved without being stuck at local minima. The solution is a near-optimal and near-feasible solution for the time-optimal problem. In the second stage, the minimum-curvature trajectory is used to warm-start the non-convex time-optimal optimization problem. In our experiments (Section~\ref{sec:planning-ablation}), we show that, by providing a well-informed starting point, the optimization process becomes more stable and efficient, improving the likelihood of finding a high-quality minimum lap-time trajectory. In particular, the two-stage planning algorithm is able to find a trajectory that is \emph{almost identical} to the best human player's trajectory. In the following subsections, the two stages are described in detail respectively.

\subsection{Minimum Curvature Trajectory Optimization}
\label{SubSec:Kap-optimal}
In Kapania et al.~\cite{kapania2016sequential}, a two-step planning algorithm based on a curvature-optimal optimization problem is proposed. It aims to approximately solve the time-optimal problem based on the hypothesis that a path with minimum curvature is a good approximation for the minimum-time trajectory. Here, we likewise formulate a curvature-optimal optimization problem \cite{kapania2016sequential}, which is in the form of a quadratic programming (QP) problem. Instead of directly adopting the same hypothesis, we derive the curvature-optimal objective from the time-optimal objective to show that curvature-optimal planning indeed approximates the time-optimal path. It justifies our practice of warm-starting the time-optimal optimization with the curvature-optimal path.

Concretely, if we assume the curvature $\kappa$ and the yaw difference angle $e_\psi$ are negligible, the time-optimal objective becomes:
\begin{equation}
    T \approx \int_{s_{0}}^{s_{f}} \frac{d s}{V_{x}} \label{Eqn:TimeObj3}
\end{equation}
The new objective is straightforward to minimize\textemdash we should maximize the velocity to minimize the lap time. Moreover, the velocity has an upper bound defined by the maximum lateral traction force and the path curvature (Eq.~\eqref{Eqn:MaxV}). It limits the vehicle speed at sharp corners with large curvature. 
\begin{equation} \label{Eqn:MaxV}
    V_{x,max} = \sqrt{a_{tire,max}/\kappa}
\end{equation}
where $a_{tire,max}$ denotes the maximum allowable acceleration generated by the tire, which is defined as $a_{tire,max}=\mu(g\cos(\theta)+a_z)$. Therefore, we can minimize the lap time by minimizing the curvature of the planned trajectory. It leads to the curvature-optimal planning problem. Under the same assumption of small curvature and yaw difference angle, we can write the curvature as a function of vehicle states: 
\begin{equation} \label{Eqn:Kap}
\kappa=\frac{d \psi}{d s}=\frac{d \psi}{d t} \frac{d t}{d s}=\frac{\dot{\psi}}{\dot{s}} \approx \frac{\dot{\psi}}{V_{x}}
\end{equation}

We can then formulate the curvature-optimal optimization problem as follows: 
\begin{align} 
\min_{\substack{\xi_1, \cdots, \xi_{N_f}, \\ u_1,\cdots, u_{N_f}}} \sum_{k=1}^{N_f} & \left(\frac{\dot{\psi}_k}{\hat{V}_{x, k}}\right)^2+w_{1} V_{y, k}^2+w_{2} (\delta_k-\delta_{k-1})^2 \label{Eqn:KapObj}\\
\mathrm{s.t.} \quad \xi_{k+1}&=A_{k} \xi_{k}+B_{k} u_{k}+D_{k} \label{Eqn:KapConstr1}\\
e_{y,k} &\in \left[W_{k, \min}, W_{k, \max}\right] \\
u_{k} &\in \left[u_{k, \min}, u_{k, \max}\right].\label{Eqn:KapConstr2}
\end{align}
Note that the approximate value of $\kappa^2$, i.e., $(\nicefrac{\dot{\psi}}{V_x})^2$, is not a quadratic function because $V_x$ is on the denominator. To derive a quadratic objective function, the common practice in curvature-optimal planning is to replace the denominator $V_x$ with a fixed speed profile $\hat{V}_x$ computed given a reference path and its curvature $\kappa$. Given $\kappa$, the optimal speed profile can be computed straightforwardly using Eq.~\eqref{Eqn:MaxV}. Therefore, we can iteratively solve the curvature-optimal planning problem by alternating between solving the optimization problem in Eq.~\eqref{Eqn:KapObj}-\eqref{Eqn:KapConstr2} and updating the optimal speed profile given the new $\kappa$. In practice, we can use the centerline of the track to initialize $\kappa$. Apart from the curvature, the objective function consists of two regularization terms to penalize large lateral velocity $V_{y, k}$ and change of steering angle $(\delta_k - \delta_{k-1})$, which are added to stabilize the QP solver. In our experiments, we choose the regularization weights as $w_1 = 0.015$ and $w_2=0.01$. To induce a QP problem, the vehicle dynamics model is also linearized regarding the solution of the last iteration to derive the \blue{inequality constraint} in Eq.~\eqref{Eqn:KapConstr2}. 

\subsection{Minimum Lap Time Trajectory Optimization} The time-optimal planning problem aims to minimize the lap time of the planned trajectory. To formulate time-optimal planning as an optimization problem, the implicit \emph{time} objective needs to be expressed explicitly as a function of vehicle states. For the time-optimal planning of racing vehicles, we are able to derive such an explicit expression using the track geometry. Following prior works~\cite{metz1989near, verschueren2014towards}, we convert the lap time, $T$, from an integral in the temporal domain to an integral in the spatial domain by converting the vehicle states to Frenet–Serret coordinates with respect to the centerline of the track:  
\begin{align}
    T= & \int_{0}^{T} 1 d t \label{Eqn:TimeObj0}
    \\
    = & \int_{s_{0}}^{s_{f}} \frac{d t}{d s} ds=\int_{s_{0}}^{s_{f}} \frac{1}{\dot{s}} ds \label{Eqn:TimeObj1} 
    \\
    = & \int_{s_{0}}^{s_{f}} \frac{1-\kappa(s) e_{y}(s)}{V_{x}(s) \cos \left(e_{\psi}(s)\right)-V_{y}(s) \sin \left(e_{\psi}(s)\right)} ds.\label{Eqn:TimeObj2}
\end{align}
Here $s_0$ and $s_f$ denote the curve length of the starting and ending points along the centerline of the track. $\kappa(s)$ denotes the curvature of the centerline curve and $V_x(s), V_y(s), e_\psi(s), e_y(s)$ denote vehicle states at the waypoint with traveling distance $s$.


We discretize the integral with respect to waypoints along the centerline of the track and arrive at the following optimization problem with respect to states $\xi$ and controls $u$: 
\begin{align} 
\min_{\substack{\xi_1, \cdots, \xi_{N_f}, \\ u_1,\cdots, u_{N_f}}} \sum_{k=1}^{N_f} & \frac{\left(1-\kappa_k \ e_{y, k}\right) \Delta s_k}{V_{x, k} \cos \left(e_{\psi, k}\right)-V_{y, k} \sin \left(e_{\psi, k}\right)} \nonumber \\
& \blue{+w_{\dot{\delta}} (\delta_k-\delta_{k-1})^2 }
\label{TimeObjDis} \\
\mathrm{s.t.} \quad \xi_{k+1}&=f(\xi_k, u_k), \label{Eqn:TimeConstr0}\\
e_{y,k} &\in \left[W_{k, \min}, W_{k, \max}\right] \label{Eqn:TimeConstr1} \\
u_{k} &\in \left[u_{k, \min}, u_{k, \max}\right], \label{Eqn:TimeConstr2}
\end{align} 
\blue{where the $(\delta_k-\delta_{k-1})$ term regularizes the rate of steering.}

The equality constraint~\eqref{Eqn:TimeConstr0} represents the nonlinear vehicle dynamics in Frenet-Serret coordinates, which we derive from the vehicle dynamics in the time domain using the same trick as in Eq.~\eqref{Eqn:TimeObj1}. Denote the vehicle dynamic model in the time domain (Section \ref{Sec:VehicleModel}) as    $\dot{\xi}=g(\xi, u)$, then the dynamics model in the spatial domain is
\begin{align}
    {\xi}'(s) &= \frac{d\xi}{ds} = \frac{d\xi}{dt}\frac{dt}{ds}=\frac{1}{\dot{s}}\dot{\xi} \nonumber \\
    &= \frac{1-\kappa(s) e_{y}(s)}{V_{x}(s) \cos \left(e_{\psi}(s)\right)-V_{y}(s) \sin \left(e_{\psi}(s)\right)} g\left(\xi(s), u(s)\right), \nonumber
\end{align}
and \eqref{Eqn:TimeConstr0}  is then obtained after discretization. 

The inequality constraints in Eq.~\eqref{Eqn:TimeConstr1} and~\eqref{Eqn:TimeConstr2} represent the track width boundary limits and the control input limits. However, in~\eqref{Eqn:TimeConstr2}, the limits of longitudinal acceleration $a_x$ depend on the vehicle state. To ensure that the acceleration commands are indeed feasible for the vehicle to execute, we iteratively solve the time-optimal optimization problem. In each iteration, we update the acceleration bounds based on the solved trajectory from the last iteration. To estimate the acceleration bounds given nominal states, we develop an empirical powertrain and longitudinal tire force model. We omit the details here and refer the readers to Appendix~\ref{Appendix: long_force}.

\color{black}


\section{Tracking Controller}
\label{Sec:Controller}
In this section, we devise the MPC-based tracking controller, which controls the vehicle to track the planned time-optimal reference trajectory. Commonly, the vehicle control problem is decoupled into longitudinal and lateral controls for simplicity\cite{kong2015kinematic,xu2018zero,tang2019disturbance}. The longitudinal controller generates pedal commands to track the reference longitudinal velocity based on the longitudinal dynamics. In contrast, the lateral controller generates steering commands to track the reference path based on the lateral dynamics. However, we want to emphasize that the longitudinal and lateral motion should be jointly optimized in the racing problem because the longitudinal and lateral dynamics of race cars are tightly coupled. As shown in Eq.~\eqref{Eqn:long_model}-\eqref{Eqn:lat_model1}, the longitudinal velocity contributes to the lateral acceleration and vice versa. Their effects cannot be neglected when the vehicle is operated at high speed and is steered swiftly in sharp corners. In addition, the weight transfer caused by large longitudinal accelerations influences the wheel load and lateral tire forces. As a result, decoupling the longitudinal and lateral dynamics leads to large prediction errors and sub-optimal control commands. To this end, we design an MPC-based tracking controller that jointly optimizes the pedal and steering commands based on the coupled longitudinal and lateral vehicle dynamics model. 

\subsection{MPC Formulation}
At each time step, we compute the nearest point on the reference trajectory to the current position of the vehicle as the reference point \blue{to calculate the tracking errors $e_\psi$ and $e_y$ in the Frenet coordinate}. We then select the $N_p$ consecutive points starting from the reference point on the reference trajectory, denoted as $\xi_k^{ref}$, $k=0, 1, ..., N_p$. The MPC controller then controls the vehicle to minimize the tracking errors relative to $\xi_k^{ref}$ over the control horizon $N_p$. The optimization problem with respect to states and controls (\ref{Eqn:states_and_controls}) is formulated as:
\begin{align} 
\min_{\substack{\xi_1, \cdots, \xi_{N_p}, \\ u_1,\cdots, u_{N_p-1}}} &
\sum_{k=1}^{N_{p}} \|\xi_{k}-\xi_{k}^{\mathrm{ref}}\|^{2}_{w_\xi}+w_{\dot{\delta}} (\delta_k-\delta_{k-1})^2 \label{Eqn:MPCObj} \\
\mathrm{s.t.} \quad \xi_{k+1}&=A_{k} \xi_{k}+B_{k} u_{k}+D_{k} \label{Eqn:MPCConstr1}\\
e_{y,k} &\in \left[W_{k, \min}, W_{k, \max}\right] \label{Eqn:MPCConstr2}\\
u_{k} &\in \left[u_{k, \min}, u_{k,\max}\right]. \label{Eqn:MPCConstr3}
\end{align}

The first term of \eqref{Eqn:MPCObj} is the weighted sum of the tracking errors with respect to the reference trajectory. The second term regularizes the changing value of the steering angle to prevent high-frequency oscillation. Eq.~\eqref{Eqn:MPCConstr1} is an equality constraint to enforce the coupled longitudinal and lateral vehicle model, that is linearized from. Solutions for nonlinear MPC have been proposed -- such as iterative least quadratic regulation (iLQR) ~\cite{chen2019autonomous}, where linearized approximations are iteratively solved from the solution in the last iteration. However, the convergence of iLQR to the nonlinear MPC requires significant computational time, and often only a few iterations are performed during control. 
Instead, we employ a linearization technique that linearizes Eq.~\eqref{Eqn:epsi}-\eqref{Eqn:lat_model2} with respect to the states and control inputs using a good initial guess for the solution, which we term the heuristic trajectory.

For the heuristic trajectory, the nearest section of the reference trajectory itself is a good candidate; however, if the car is significantly deviated from the reference it may lead to linearization errors at the beginning of the control horizon. We design the heuristic trajectory, $\hat{\xi}_{k}$,  that smoothly transitions over the control horizon from the current state to the reference:
\begin{equation} \label{Eqn:refstate}
    \hat{\xi}_k = \xi_{k}^{ref} + \lambda_{k}(\xi_0 - \xi_{0}^{ref}),
\end{equation}
where $\xi_0$ denotes the current state of the vehicle, and $\xi_{k}^{ref}$ denotes the reference states at the $k^\text{th}$ step in the horizon. The weight sequence $\{\lambda_k\}_{k=0}^{N_p}$ is a monotonically decreasing sequence with $\lambda_0=1$ and $\lambda_{N_p}=0$. For simplicity, we adopt a linear transition: $\{\lambda_k\}_{k=0}^{N_p} = \{0, 1, ..., N_p\} / N_p$. Thus we linearize Eq.~\eqref{Eqn:MPCConstr1} with respect to the heuristic trajectory that gradually transitions from the current state to the reference over the control horizon $N_p$. In experiments, we found this led to better performance than using the reference trajectory itself as the heuristic trajectory.

Similar to the case in planning, the limits of longitudinal tire forces $a_x$ in Eq.~\eqref{Eqn:MPCConstr3} are state-dependent. We used the same empirical model to compute the acceleration bounds given a nominal state vector.  Instead of iteratively updating the nominal states, we simply choose the nominal states as the ones defined in Eq.~\eqref{Eqn:refstate}. In the controller, we also need to convert the calculated longitudinal tire acceleration $a_x$ to actual pedal commands (i.e., throttle and braking) for execution in GTS. We rely on the same empirical powertrain and longitudinal tire model to compute the pedal commands. More details are given in Appendix~\ref{Appendix: long_force}.

\subsection{Hyperparameter Optimization} \label{Subsec:weight-tuning}
The objective function in Eq.~\eqref{Eqn:MPCObj} is a weighted sum of three tracking errors. In practice, we must carefully tune the weights to achieve the best tracking performance. To avoid tedious and heuristic manual tuning, we adopt the auto-tuning algorithm Covariance Matrix Adaptation Evolution Strategy (CMA-ES) \cite{hansen2016cma} to automate the weight-tuning process.  In particular, in Section \ref{Sec:ExpResult}, we present an ablation study of different controller configurations. For each configuration, we tune the controller weights using the following strategy; this ensures that each configuration is compared fairly without the need for hand-tuning ideal weights.

The CMA-ES algorithm searches for the MPC weights, which minimize a heuristic objective function. It minimizes the objective function by iteratively sampling MPC weights, running the corresponding controller in the simulator, and then updating the weights based on the closed-loop trajectories. We design the following objective function for our experiments: 
\begin{align} \label{Eqn:CMA}
\min _{\substack{w_{V_x}, w_{V_y}, w_\omega \\ w_{e_{\psi^{\prime}}}, w_{e_y}, w_{\dot{\delta}}}} & \left|e_{y, \max }\right| +\sum_{k=1}^{N_f} \frac{1}{N_f}\left[\left|e_{y, k}\right|+\left|V_{x, k}-V_{x, k}^{ref}\right|\right] \nonumber \\
&  +w_t t_{c, k}- w_s \Delta s_k
\end{align}
where $\left|e_{y, \max }\right|$ is the maximum lateral distance error, and $\left|e_{y, k}\right|$ and $\left|V_{x, k}-V_{x, k}^{ref}\right|$ penalize the average errors of lateral distance and longitudinal velocities with respect to the reference velocity $V_{x, k}^{ref}$. The last two terms $t_{c,k}$ and $\Delta s_k$ denote \blue{collision time with track boundary}, and distance traveled at each time. With proper weights, the last two terms penalize collision and encourage longer traveling distances per time step, respectively. Two weights $w_t=10$ and $w_s=0.001$. The variable $N_f$ denotes the total number of time steps.

\section{Experiments}
\label{Sec:ExpResult}

\renewcommand\arraystretch{2.0} 
\begin{table*}[t]
\caption{Planning and control experimental results} \label{Tab:results}
    \centering
\begin{threeparttable}
    \begin{tabular}{l|cccc}
    \hline \hline

        Experiments $^*$ &
        \makecell[c]{Planning \\ lap time\ ($s$)} &
        \makecell[c]{Closed-loop \\ lap time\tnote{1} \ ($s$)} & \makecell[c]{Longitudinal velocity \\ MAE\tnote{2} \ ($m/s$) }  & \makecell[c]{Lateral distance \\ MAE\tnote{2} \ ($m$) }    \\
    \hline
    
        Main experiment (Tokyo Expressway) & 99.48 &  101.03 $\pm$ 0.089 (100.80) & 0.308 $\pm$ 0.363 (1.872) & 0.158 $\pm$ 0.340 (1.508) \\

        \xmark \hspace{2pt} Wind drag force &  96.89 &  101.86 $\pm$ 0.056 (101.80) & 2.13 $\pm$ 1.535 (7.362)  & 0.292 $\pm$ 0.715 (4.800) \\
        
        \xmark \hspace{2pt} Longitudinal weight transfer &  100.36 & 101.52 $\pm$ 0.060 (101.43) & 0.300 $\pm$ 0.413 (2.218)  & 0.168 $\pm$ 0.368 (1.682) \\

        \blue{\checkmark \hspace{2pt} Lateral weight transfer} & 99.52 & 100.94 $\pm$ 0.119 (100.86) & 0.457 $\pm$ 0.350 (1.524)   & 0.143 $\pm$  0.310 (1.619) \\

        \xmark \hspace{2pt} Time-optimal trajectory &  101.28 & 103.80 $\pm$ 0.207 (103.58) & 0.925 $\pm$ 0.980 (5.743)  & 0.180 $\pm$ 0.527 (4.246) \\

        \xmark \hspace{2pt} MPC with coupled model &  99.48 &  103.48 $\pm$ 0.485 (102.86) & 1.312 $\pm$ 1.456 (7.071)  & 0.377 $\pm$ 0.736 (4.330) \\

        \blue{\xmark \hspace{2pt} Smooth reference transition} & 99.48 & 103.56 $\pm$ 0.118 (103.28) & 1.691 $\pm$ 1.905 (7.887)  & 0.258 $\pm$ 0.580 (3.540) \\

        \hline

        \blue{Sharper track experiment (Autopolis)} & 93.00 & 96.623 $\pm$ 0.103 (96.48) & 1.629 $\pm$ 0.591 (3.374)  & 0.228 $\pm$ 0.410 (2.385) \\

        \blue{\checkmark \hspace{2pt} Lateral weight transfer } & 92.89 & 96.556 $\pm$ 0.083 (96.44) & 1.582 $\pm$ 0.556 (3.210)  & 0.230 $\pm$ 0.409 (2.375) \\
         
    \hline \hline
    \end{tabular}
    
\begin{tablenotes}
    \item[$*$] The main experiment is conducted on the Tokyo Expressway. We then conduct ablation studies that remove (\xmark) or add (\checkmark) one feature to the main experiment. Finally, we transfer the main control system to a new track Autopolis with sharper corners.
    \item[1] The closed-loop lap time is presented in the format of $\mathrm{mean} \pm \mathrm{std} \ (\mathrm{minimum})$. The statistics are computed over 10 trials. 
    \item[2] MAE stands for mean absolute error. The errors are presented in the format of $\mathrm{mean} \pm \mathrm{std} \ (\mathrm{maximum})$. The statistics are computed over 10 trials.
\end{tablenotes}
\end{threeparttable}
\end{table*}

In this section, we conduct comprehensive experiments to evaluate the proposed racing framework and challenge the best human racing performance. In Sec.~\ref{sec:setup}, we summarize the setup of the experiments. In Sec.~\ref{sec:main-result}, we present the results of the main experiment, where we test the racing framework with all the proposed elements. The objective of the main experiment is to see if we can achieve human expert-level performance with the proposed racing framework. From Sec.~\ref{sec:model-ablation} to \ref{sec:control-ablation}, we conduct a series of ablation studies to validate the effectiveness of the proposed improvements. Specifically, we aim to answer the following questions in the ablation studies:

\begin{itemize}
    \item Is it necessary to model the aerodynamic force and weight transfer to achieve the shortest lap time?
    \item Does the two-stage time-optimal planning framework outperform curvature-optimal planning and generate a racing line close to the human-best racing line? 
    \item Does the coupled MPC controller achieve better tracking performance than the decoupled controllers?
\end{itemize}

\subsection{Experiment Setup} \label{sec:setup}
The closed-loop experiments were conducted in the high-fidelity racing simulator Gran Turismo Sport~\cite{GTS_web}. We employ the same method as prior researchers to communicate with the PS4 \cite{fuchs2021super}, although in our case we use only one PS4 and one car on the track at a time. The vehicle we tested on is the Mazda Demio XD Turing '15, and the racing track is the Tokyo Expressway Central Outer Loop. We identified the vehicle parameters based on data collected with a simple nominal controller. The parameters identified are summarized in Table~\ref{tab:Parameters}. In addition, we have access to human racing trajectories at different levels, including the trajectories from the best human racers~\cite{fuchs2021super} for comparison. 

\renewcommand\arraystretch{1.5} 
\begin{table}[b]
    \centering
    \caption{Parameters of Mazda Demio XD Turing '15 in GTS}
    \begin{tabular}{lc}
    \hline \hline
    Parameter & Value \\
    \hline
    Total mass $m$ &  1355.2$kg$  \\
    Length from CoG to front wheel $l_f$ & 0.9338$m$ \\
    Length from CoG to rear wheel $l_r$ & 1.6363$m$ \\
    Width of chassis & 2.008$m$ \\
    Height of CoG $h_c$ & 0.6161$m$ \\
    Friction ratio $\mu$ & 1.25 \\
    Wind drag coefficient $C_{xw}$ & 0.1302$kg/m$ \\
    Moment of inertia $I_{zz}$ & 2475.33$Nm$ \\
    \hline \hline
    \end{tabular}
    \label{tab:Parameters}
\end{table}

We conducted the experiments on a Dell G7 Laptop with an Intel Core i7-9750H CPU and 16GB Memory. A PlayStation 4 Pro is connected to the computer via a wired router to run GTS, \blue{where the control frequency is 10Hz.} We use cvxpy~\cite{diamond2016cvxpy} and qpsolver~\cite{domahidi2013ecos} to solve the QP problem in MPC, and we use Cadasi~\cite{Andersson2019, wachter2006implementation} to solve the nonlinear programming problems in time-optimal planning. We provide details of experiments in Gran Turismo Sport in Appendix~\ref{Appendix: GTS}.

\subsection{Main Experimental Results} \label{sec:main-result}
\label{SubSec:performance}

\begin{figure*}[t]
    \centering
    \includegraphics[width = \textwidth]{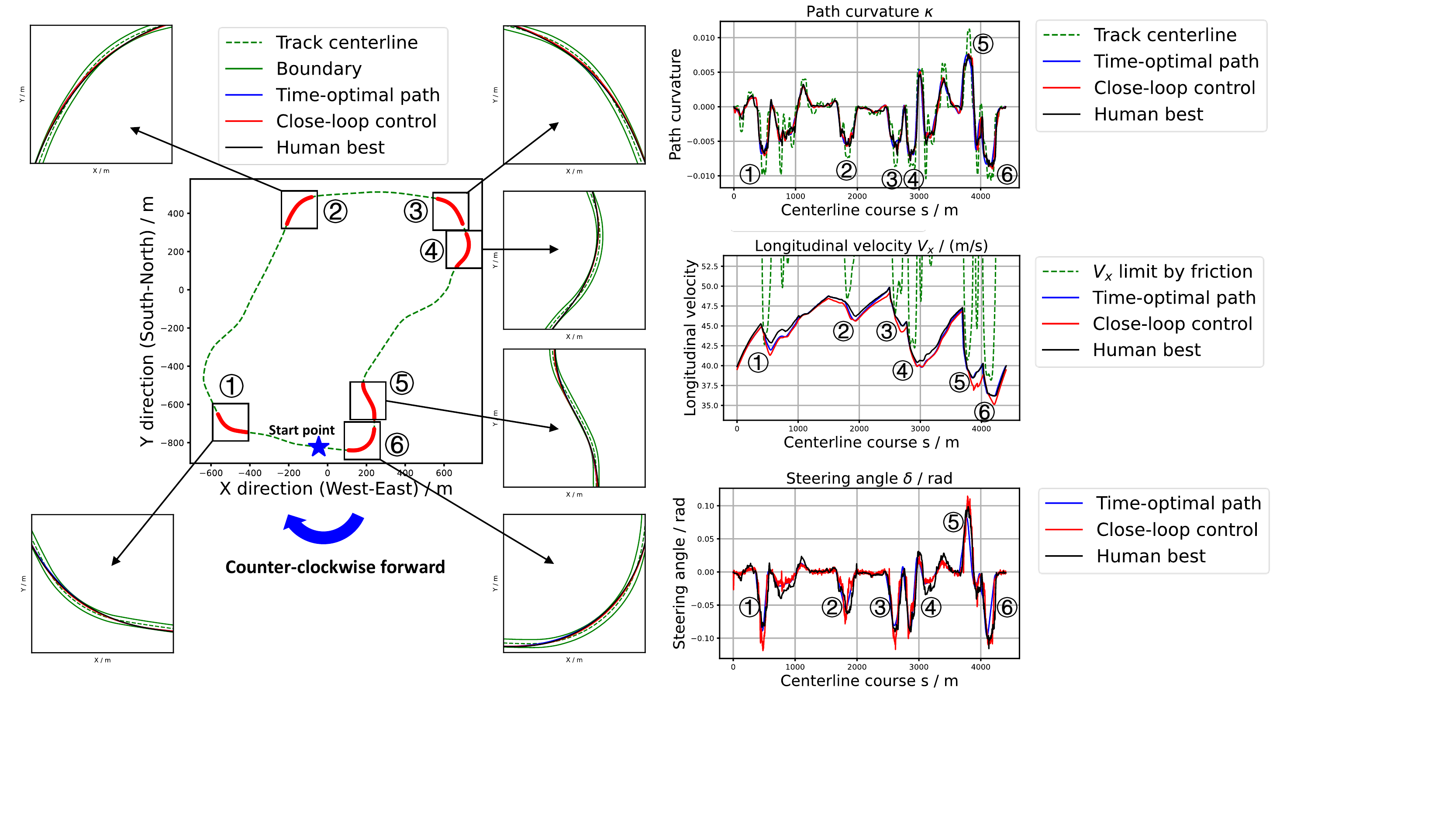} 
    \caption{\blue{ \textbf{Tokyo Expressway Central Outer Loop}. Planning and closed-loop control results and comparison with the human-best trajectory. \textbf{Left}: The overview of the racing track and all the paths at six crucial corners marked by \textcircled{1} - \textcircled{6}. \textbf{Right}: three figures show the path curvature, longitudinal velocity, and steering angle of the planned time-optimal, close-loop, and human-best trajectories.} }
    \label{fig:main_experiment}
\end{figure*}
\blue{
In the main experiment, we evaluated the racing framework with all the proposed elements. In summary, 1) we model the wind drag force and longitudinal weight transfer\textemdash excluding lateral weight transfer\textemdash in the bicycle vehicle dynamics model used for planning and control (Sec.~\ref{Sec:VehicleModel}); 2) we use the two-stage time-optimal planning algorithm (Sec.~\ref{Sec:TrajPlan}) to generate the reference trajectory; 3) we use the MPC controller which jointly optimizes the pedal and steering command based on the coupled longitudinal and lateral vehicle model (Sec.~\ref{Sec:Controller}) to control the race car to track the smoothed planned reference trajectory transitioning from the current state to the predicted trajectory. } For the planning experiment, the planning algorithm outputs the same planned trajectories consistently over different trials, so we only report the result of a single trial. For the closed-loop control experiments, we controlled the race car to track the reference trajectory and applied the hyperparameter tuning method introduced in Sec.~\ref{Subsec:weight-tuning} to tune the MPC weights. Then we sampled the closed-loop trajectory tracking the same reference trajectory 10 times and reported the statistics over the 10 trials to account for the uncertainty in the simulator.

The results are shown in Table~\ref{Tab:results} and Fig.~\ref{fig:main_experiment}. In Fig.~\ref{fig:main_experiment}, the left figure consists of an overview of the racing tack and zoomed-in figures showing the trajectories in 6 crucial corners with large curvature. We mark the crucial corners with \textcircled{1} - \textcircled{6}. Those crucial corners put high demands on both planning and control, as the race car needs to carefully control its speed and race line inside the corners to pass the corners fast without losing traction. On the right-hand side, we plot the curvature, longitudinal velocity, and steering angle profiles of the time-optimal and closed-loop paths and compare them with the human-best racing recording. 

We first compare the time-optimal path with the human-best path. One of the most important features of human racers' racing line in corners is the \emph{out-in-out principle}~\cite{betz2022autonomous}, where the race car gets close to the \emph{outer} boundary of the track at the entrance of the corner, then turn to the \emph{inner} boundary at the apex (i.e., location with the maximum curvature), and finally exit the corner along the outer boundary. This special racing line can significantly reduce the maximum curvature at the sharp corner and hence enhance the velocity limits. As shown in Fig.~\ref{fig:main_experiment}, the human-best path follows the out-in-out racing line and obtains smaller curvature at crucial corners than the track centerline. Similarly, the time-optimal path has almost identical racing lines and curvature as the human-best path. In addition, the time-optimal planning algorithm can also accurately estimate the velocity limits given the path curvature, enabling the race car to pass the corners with the maximum allowable speed as the best human racer does.

We then examine whether the controller can track the planned time-optimal trajectory well. As shown in Table~\ref{Tab:results}, the average absolute longitudinal velocity error is $0.308$ $m/s$, which is less than $0.7\%$ of the average driving speed. Moreover, the average absolute lateral distance error is $0.158 m$, which remains low over the whole track. In particular, the race car can precisely track the out-in-out racing lines inside the crucial corners. \blue{One difference between the time-optimal reference path and the closed-loop path is the steering angle. The actual steering angle is larger than the reference at corner \textcircled{1} and \textcircled{3} to provide more lateral forces at apex points. 
In addition, the actual steering at corner \textcircled{5} and \textcircled{6} is closer to the human best than the planned path.}

\begin{figure}[t]
    \centering
    \includegraphics[width=1.0\columnwidth]{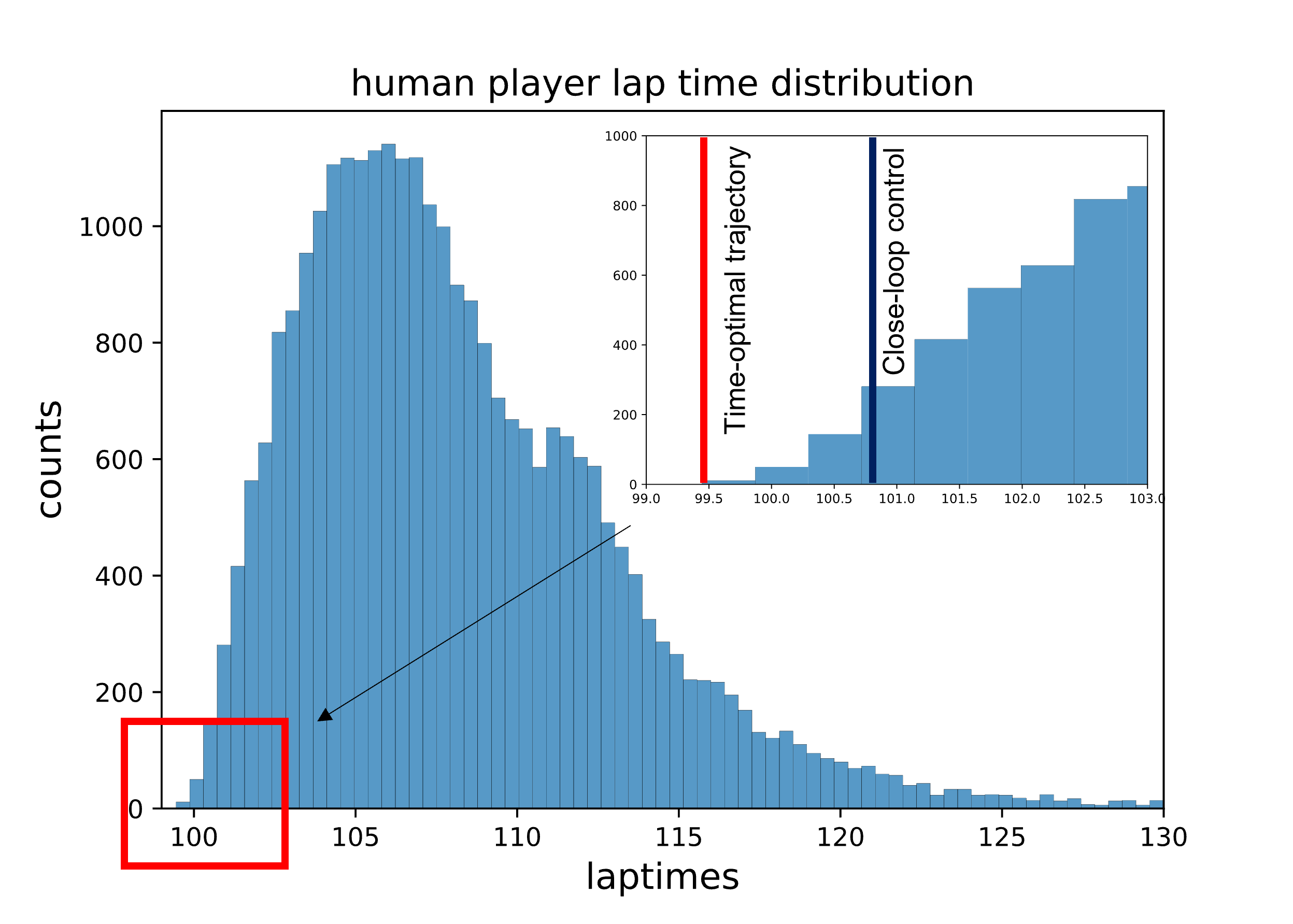}
    \caption{Lap time distribution of human players for a time-trial race competition on Tokyo Track with Demio vehicle (see Appendix \ref{Appendix: GTS} for more details). The best human lap time is $99.43s$; the best lap time of the time-optimal trajectory is $99.48s$ (top $0.35\%$), and the best lap time of the closed-loop trajectory is $100.83s$ (top $0.95\%$).}
    \label{fig:laptime_distribution}
\end{figure}

Lastly, we summarize the lap times of the planned and the closed-loop trajectories in Table~\ref{Tab:results} and compare the lap times with the human racers' recordings in Fig.~\ref{fig:laptime_distribution}. The lap time of the time-optimal trajectory (i.e., $99.48s$) is comparable to the human-best recording (i.e., $99.43s$). Its performance is within the top $0.35\%$ of all the human racers. Additionally, the best closed-loop lap time (i.e., $100.83s$) is within the top $0.95\%$ of human racers' performance. The results show that our control system is able to achieve human expert-level performance.

\subsection{Vehicle Model Analysis}\label{sec:model-ablation}
\begin{figure}[t]
    \centering
    \includegraphics[width=0.9\columnwidth]{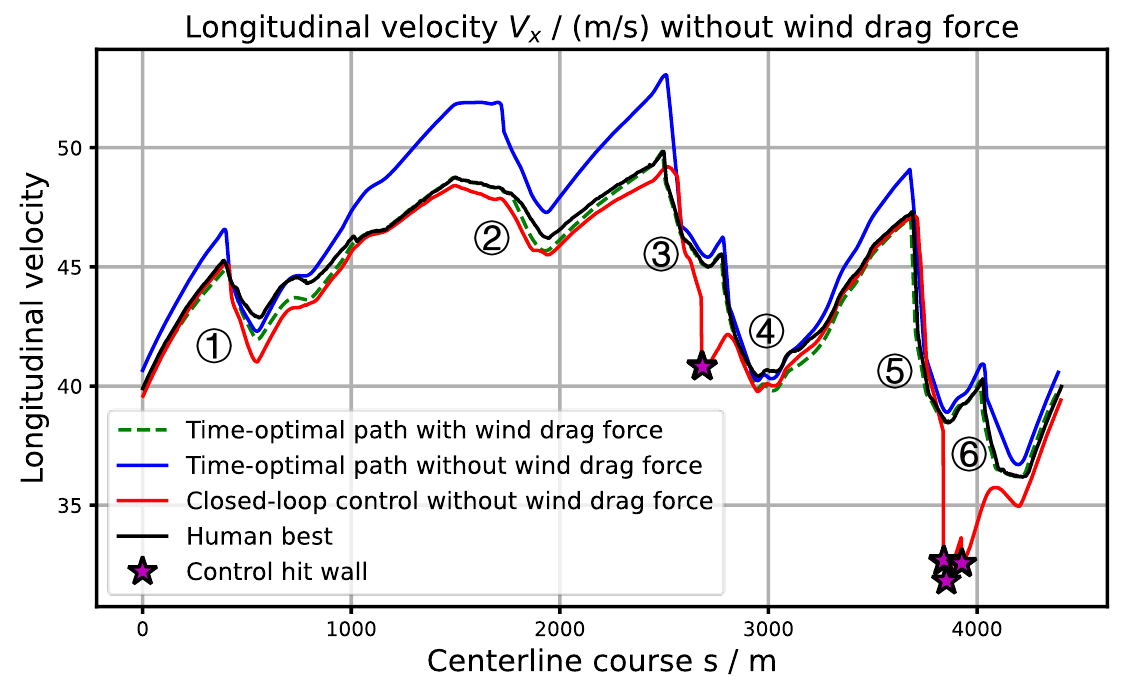}
    \caption{\blue{ Longitudinal velocity of the time-optimal and closed-loop trajectories while removing the wind drag force in the model ($C_{xw}=0$). } }
    \label{fig:ablation_Cxw}
\end{figure}

\begin{figure}[t]
    \centering
    \includegraphics[width=0.9\columnwidth]{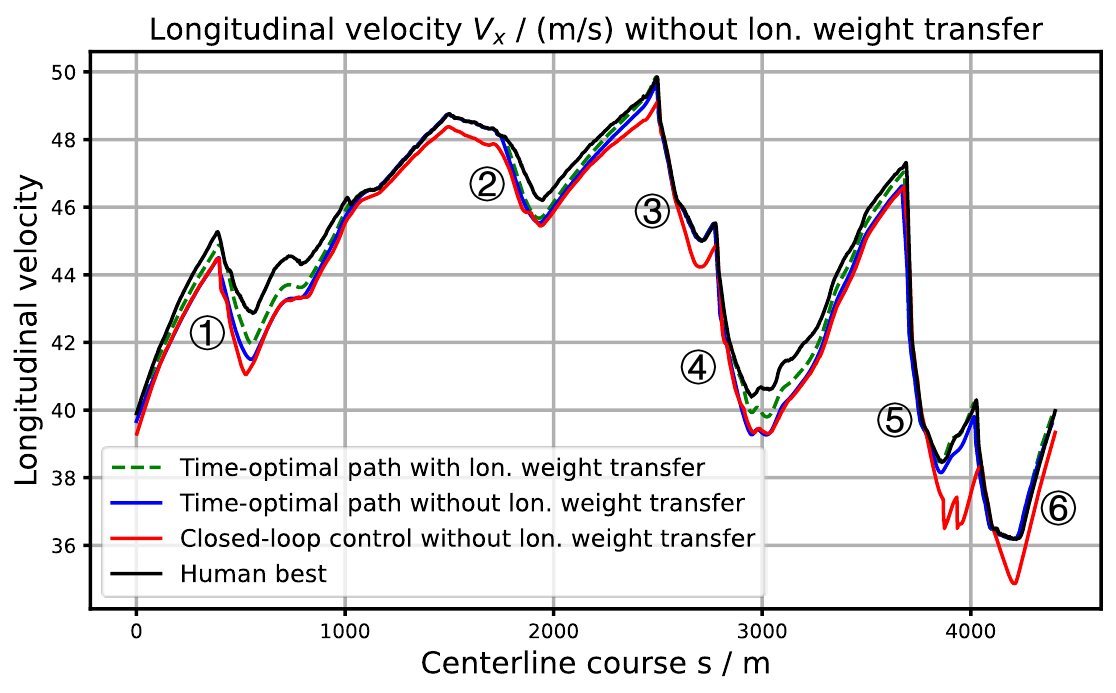}
    \caption{\blue {Longitudinal velocity of the time-optimal and closed-loop trajectories while removing the longitudinal weight transfer in the model ($\Delta W =0$). }}
    \label{fig:ablation_hc}
\end{figure}

\begin{figure}[t]
    \centering
    \includegraphics[width=0.9\columnwidth]{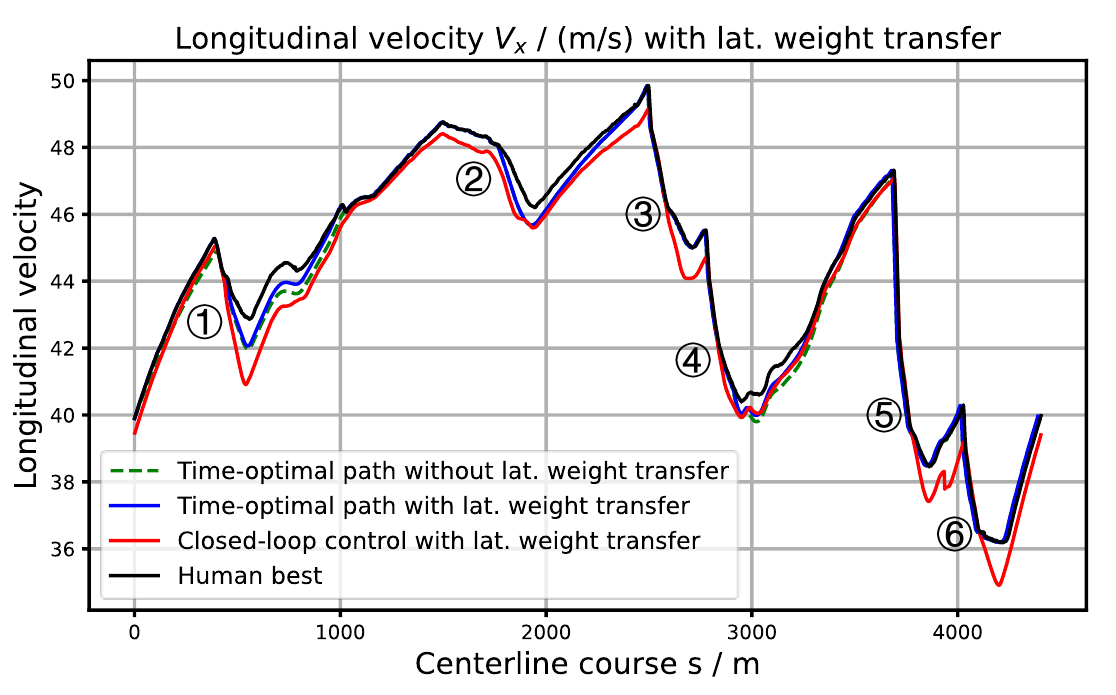}
    \caption{\blue{ Longitudinal velocity of the time-optimal and closed-loop trajectories while adding the lateral weight transfer in the double-track vehicle model. }}
    \label{fig:ablation_lateral}
\end{figure}

In this subsection, we validate the necessity of modeling the wind drag force, longitudinal and lateral weight transfer in the vehicle model for both planning and control. \blue{We conducted three ablation studies where in the first two experiments, the wind drag force and the longitudinal weight transfer effect were removed, respectively, achieved by setting the wind drag coefficient $C_{xw}=0$ and longitudinal weight transfer $\Delta W = 0$, respectively; in the last experiment, we include the lateral weight transfer in the double-track vehicle model (Appendix~\ref{Appendix: 4wheel model})}. We show the longitudinal velocity profiles of the planned and closed-loop trajectories in Fig.~\ref{fig:ablation_Cxw},  \ref{fig:ablation_hc} and \ref{fig:ablation_lateral}. The lap times and tracking errors are summarized in the second to fourth rows of Table~\ref{Tab:results}. 

As shown in Eq.~\eqref{Eqn:long_model} and \eqref{Eqn:Fxw}, the wind drag force exerted on the longitudinal direction damps the longitudinal acceleration. Without accounting for the wind drag force, the planner over-anticipates the acceleration ability of the race car, resulting in an overly higher speed profile than with wind drag force (green curve in Fig.~\ref{fig:ablation_Cxw}) and an unrealistic lap time of $96.89s$. As a result, the reference trajectory cannot be followed by the tracking controller online. The modeling error also limits the performance of the MPC controller, resulting in large tracking errors as shown in Table~\ref{Tab:results}. It also leads to critical collisions at the corner \textcircled{3} and \textcircled{5} where the vehicle speed significantly dropped. As a result, the best closed-loop lap time ($101.80s$) is much worse than the best lap time achieved when the wind drag force is considered ($100.83s$).

Additionally, removing the longitudinal weight transfer effect mainly affects the trajectory planner at the sharp corners, where the weight transfer effect redistributes the wheel loads between the front and rear wheels. Without considering the longitudinal weight transfer, the planner falsely estimates the tire forces in the sharp corners. As a result, we observe that the exiting velocities of the planned trajectories after the corner \textcircled{1} and \textcircled{4} (Fig.~\ref{fig:ablation_hc}) are much lower than the planned trajectory in the main experiment (Fig.~\ref{fig:main_experiment}). It results in a much slower planned lap time ($101.36s$) and closed-loop lap time ($101.43s$). 

\blue{
We also conducted an ablation study on the vehicle lateral dynamics. We replace the vehicle bicycle model in the main experiment (Fig.~\ref{fig:main_experiment}) with the vehicle double-track model with lateral weight transfer in both planning and control; however, it has fewer influences on the lateral dynamics. The planned and closed-loop velocity in Fig.~\ref{fig:ablation_lateral} is similar to the main experiment (Fig.~\ref{fig:main_experiment}). In addition, the ablation of lateral weight transfer does not affect the lap time of planning ($99.52s$) and control ($100.86s$). Consequently, lateral weight transfer does not contribute much in our racing environment. We analyze the root cause of these results in Appendix~\ref{Appendix: 4wheel model}.
}
 
\subsection{Planning Algorithm Analysis}\label{sec:planning-ablation}
\begin{figure}[t]
    \centering
    \includegraphics[width=0.90\columnwidth]{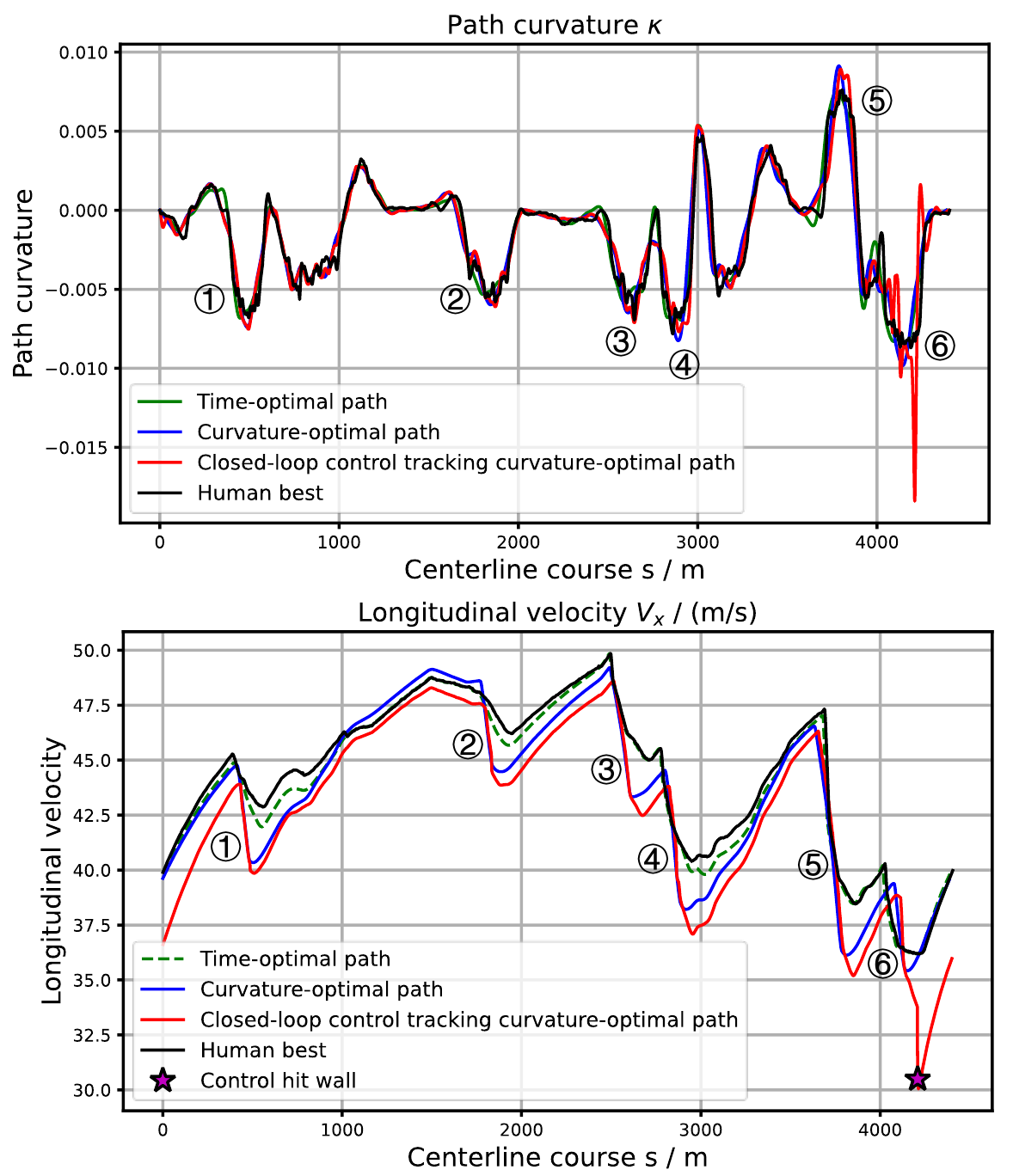}
    \caption{\blue{ The curvature-optimal trajectory and the tracking experiment. \textbf{Upper}: curvature of path; \textbf{Lower}: longitudinal velocity profile.} }
    \label{fig:track_mint}
\end{figure}

In this subsection, we conduct two ablation studies to analyze the two-stage time-optimal planning framework. The first question we are curious about is \emph{whether the proposed planning algorithm can indeed find a faster trajectory than the curvature-optimal one}. We plot the experimental results using curvature-optimal planning in Fig.~\ref{fig:track_mint}. The curvature-optimal path is similar to the time-optimal solution, but it has larger curvature at some crucial corners (e.g., \textcircled{4} and \textcircled{5}) than the human-best trajectory. \blue{This is because curvature-optimal optimization approximates the objective and dynamics model, which brings numerical errors to the solution.} As a result, the vehicle has lower speed limits in these crucial corners, as shown in the lower part of Fig.~\ref{fig:track_mint}. As shown in Table~\ref{Tab:results}, the planning lap time ($101.28s$) is, therefore, much worse than the time-optimal trajectory. Furthermore, the curvature-optimal trajectory cannot be tracked well by the MPC controller. A larger time gap between the planned trajectory and the closed-loop trajectory is observed. The tracking error of the longitudinal velocity is particularly larger. In particular, a critical collision occurs at the corner \textcircled{6}, leading to a maximum velocity error of up to $5.7 m/s$. The results show that while the curvature-optimal QP problem is easy to solve, the indirect objective function and the modeling error caused by linearization make the solution neither optimal nor feasible to the original time-optimal planning problem. It verifies the necessity and performance advantage of our two-stage time-optimal planning algorithm. 

The other question we are interested in is \emph{whether warm-starting the time-optimal optimization with the curvature-optimal trajectory is indeed necessary and effective.} In our experiments, we used Casadi (IPOPT) to solve the nonlinear time-optimal problem. \blue{If we only initialize the solver with heuristic initial guesses, for example, the track centerline at low velocity ($10 m/s$), the solver fails to find a feasible solution after running 143 seconds with a maximum of 3000 iterations. In contrast, when we use the curvature-optimal trajectory to warm-start the solver, the solver is able to find a solution within 36 iterations in 0.94 seconds.} It verifies that warm-starting with the curvature-optimal trajectory is essential to ensure a stable numerical solution to the time-optimal problem and achieve a well-planned trajectory that is close to the human-best one. 

\subsection{MPC Controller Analysis}\label{sec:control-ablation}
Lastly, to validate the necessity of using the coupled longitudinal and lateral dynamics model in the MPC control, we conduct an ablation experiment of the controller. Instead of the coupled MPC controller, we track the same time-optimal trajectory with two MPC controllers decoupled from Eq.~\eqref{Eqn:MPCObj}. The longitudinal controller controls the pedal to track the reference longitudinal velocity profile while only considering the longitudinal dynamics. The lateral controller controls the steering wheel to minimize the tracking errors of the rest of the states while assuming the vehicle perfectly follows the reference longitudinal velocity. In these two controllers, the longitudinal and lateral dynamics models are linearized with respect to the reference trajectory separately. 

\begin{figure}[t]
    \centering
    \includegraphics[width=0.9\columnwidth]{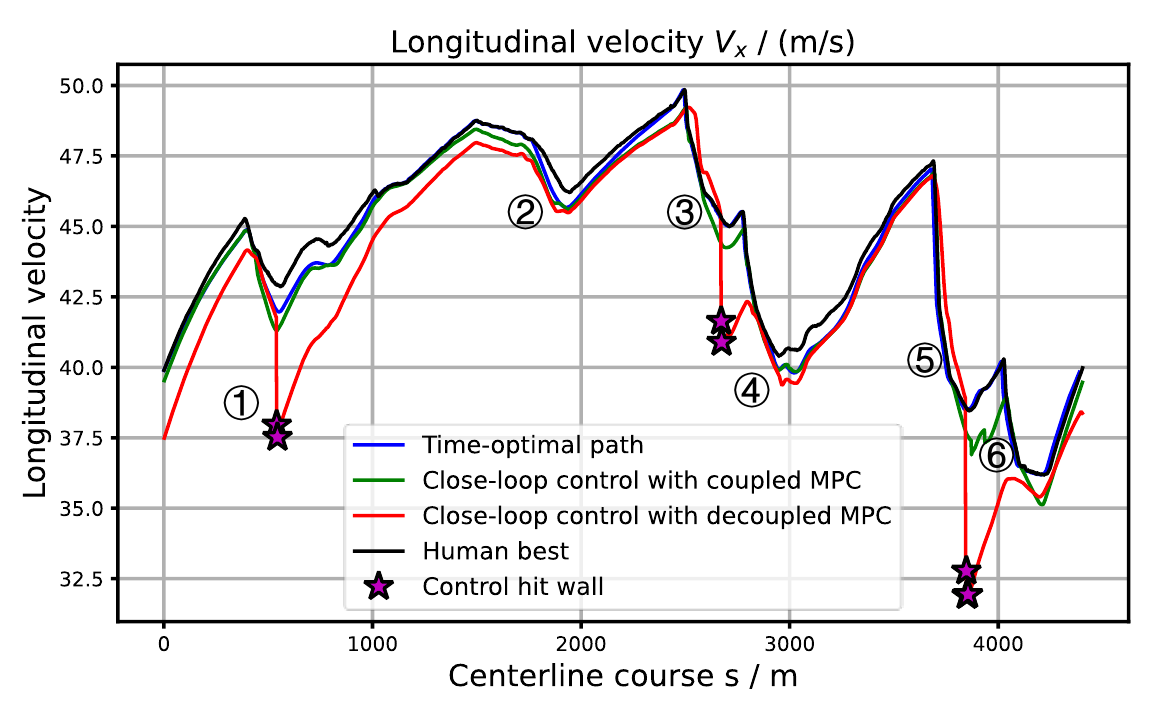}
    \caption{\blue{Closed-loop longitudinal speed profile when tracking the time-optimal trajectory using two MPC controllers controlling the longitudinal and later motions separately.}}
    \label{fig:track_dec_MPC}
\end{figure}

The closed-loop speed profile is shown in Fig.~\ref{fig:track_dec_MPC}. It can be seen that three critical collisions take place at the corner \textcircled{1}, \textcircled{3}, and \textcircled{5}. It is because the decoupled MPC controller ignores the mutual influence of the longitudinal and lateral motions in Eq.~\eqref{Eqn:long_model}-\eqref{Eqn:lat_model1}, which is especially significant under high speed and at sharp corners. As a result, the average closed-loop lap time is significantly worse ($103.48s$), as shown in Table~\ref{Tab:results}. We can therefore conclude that the coupled MPC controller is crucially important to accurately track the time-optimal trajectory and achieve minimum lap time. 

\begin{figure}
    \centering
    \includegraphics[width=0.9\columnwidth]{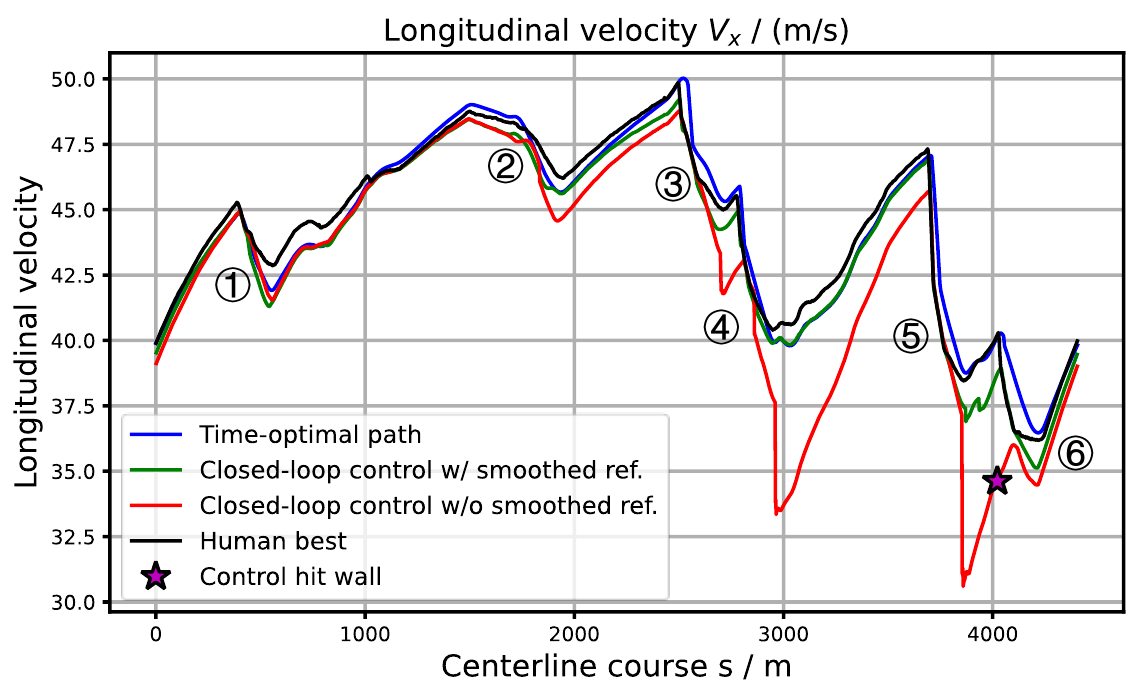}
    \caption{\blue{The closed-loop velocity of MPC with and without smoothed reference trajectory transiting from the current state to the predicted reference. }}
    \label{fig:track_nosmooth_ref}
\end{figure}

\blue{In addition, we compare the results of removing the reference trajectory transition method adopted in Eq.~\eqref{Eqn:refstate} by setting all $\lambda_k=0$. 
As shown in Fig.~\ref{fig:track_nosmooth_ref} and Table~\ref{Tab:results}, the closed-loop trajectory without smoothed reference transition has larger tracking errors in longitudinal velocity, especially at corner \textcircled{3}, \textcircled{4} and \textcircled{5}. When the car accelerates after corner \textcircled{5}, the larger tracking errors also cause a collision. This is because the MPC without smoothed reference transition attempts to minimize the lateral distance immediately, leading to abrupt maneuvers before the sharp corners. Therefore, the closed-loop lap time is worse ($103.56 s$) than using reference transition.
}

\begin{figure*}[t]
    \centering
    \includegraphics[width = 0.95\textwidth]{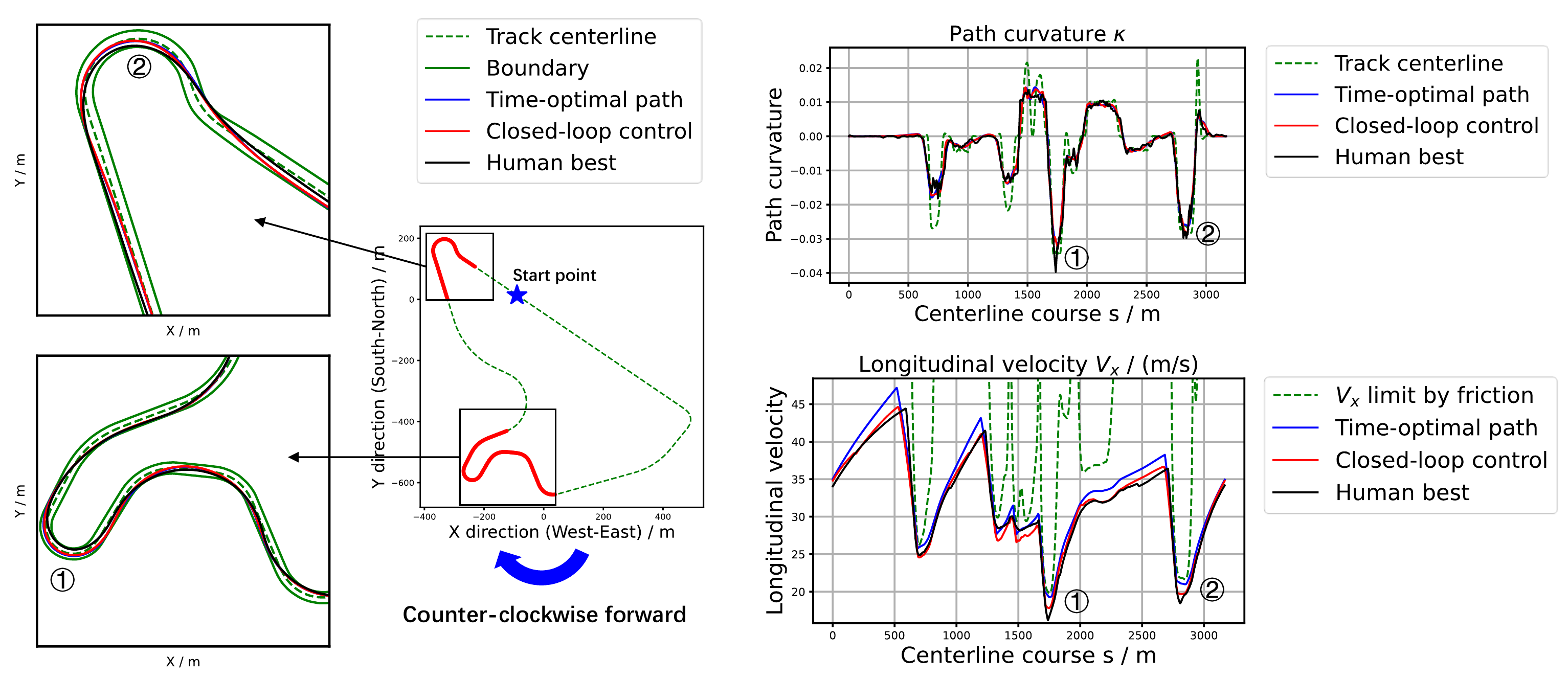} 
    \caption{\blue{ \textbf{Autopolis International Racing Course - Shortcut Course}. Planning and close-loop control results and comparison with the human-best trajectory. \textbf{Left}: The overview of the racing track and all the paths at two series of corners marked by \textcircled{1} and \textcircled{2}. \textbf{Right}: three figures show the path curvature and longitudinal velocity of the planned time-optimal, closed-loop, and human-best trajectories.} }
    \label{fig:auto}
\end{figure*}

\subsection{Sharper track} \label{Subsec: sharper_track}
In this subsection, we present the experimental results on a more challenging track with sharper corners, known as the Autopolis International Racing Course\textemdash Shortcut Course (\textbf{Autopolis}). The track layout consists of a U-shape corner at the top, marked as \textcircled{1}, and another U-shape corner at the bottom, marked as \textcircled{2}, followed by a series of sharp corners (Fig.~\ref{fig:auto}). Unlike the Tokyo Expressway Central Outer Loop (Tokyo Expressway), Autopolis does not have walls along the track boundary. Thus, the car can drive off the track to take a shorter path. In fact, as shown in Fig.~\ref{fig:auto}, the best human player indeed drives the car partially off-course to cut the apex at U-shape corners \textcircled{1} and \textcircled{2}. However, this leads to abrupt changes in tire friction outside the track, resulting in inaccuracies in the vehicle model. Therefore, we still constrain the race car from running off-course to prioritize stability and safety. Consequently, the planned time-optimal path keeps a moderate distance from the boundary. It achieves a lower average path curvature and, thus, higher velocity limits at the apex. Meanwhile, it needs to traverse a longer distance through the corners. These two distinct behaviors correspond to the traveling distance-velocity trade-off shown in Eq.~\eqref{Eqn:TimeObj3}. The human player and our autonomous racing algorithm use two different but reasonable paths to minimize the lap time. 

Regarding the lap times, the best lap time of human players is $95.40s$, while our planned trajectory yields a faster lap time of $93.00s$ (Table~\ref{Tab:results}). This discrepancy arises because the planner tends to overestimate acceleration due to the less accurate approximated linear model (Appendix~\ref{Appendix: long_force}) at higher speeds. However, the closed-loop control only practically achieves lower velocity, leading to a substantial velocity error and a closed-loop lap time of $96.48s$. Additionally, we observe that the lateral distance error on Autopolis ($0.228 m$) exceeds that of Tokyo Expressway ($0.158 m$). In conclusion, our model-based control system demonstrates effective planning strategies on the challenging Autopolis track, allowing the vehicle to navigate sharp corners while maintaining on-track stability and achieving competitive lap times.

In addition, we study the influence of lateral weight transfer on sharper corners. Similar to section~\ref{sec:model-ablation}, we adopt the double-track model (Appendix~\ref{Appendix: 4wheel model}) with lateral transfer Eq.~\eqref{Eqn: lateral weight transfer} in planning and control. The experimental results in Table~\ref{Tab:results} show that incorporating the lateral weight transfer leads to modest improvements in the planning and best closed-loop control lap time by $0.11s$ and $0.04s$. It also helps reduce the velocity tracking error but increases the lateral distance error. In total, the benefit of lateral weight transfer to the closed-loop racing performance at sharper corners is limited.
\color{black}


\begin{table}[b]
\caption{Tracking human-best experimental results} \label{Tab:track_human_results}
    \centering
\begin{threeparttable}
    \begin{tabular}{ccc}
    \hline \hline
        \makecell[c]{Closed-loop \\ lap time\ ($s$)} & \makecell[c]{Longitudinal velocity \\ MAE\ ($m/s$) }  & \makecell[c]{Lateral distance \\ MAE\ ($m$) }    \\
    \hline
    
        \makecell[c]{100.7 $\pm$ 0.292 \\ (100.54)} & 
        \makecell[c]{0.503 $\pm$ 0.520 \\ (2.058)} &
        \makecell[c]{0.074 $\pm$ 0.147 \\ (1.005)} \\
        
    \hline \hline
    \end{tabular}
\end{threeparttable}
\end{table}

\section{Discussion} \label{Sec:Discussion}
While the lap time of the time-optimal trajectory ($99.48s$) almost reaches the best human lap time ($99.43s$), there exists a time gap of $1.35s$ when we control the race car to track it. It triggers our interest in looking into this phenomenon and finding future directions for further improvements. Since the human-best trajectory was obtained by driving the same race car in the simulator, it is a feasible trajectory for the race car to track. Hypothetically, if the trajectory planner was able to find the human-best trajectory as the solution, there should exist a tracking controller that is able to perfectly track the reference and reproduce the human-best lap time. 

\begin{figure}[t]
    \centering
    \includegraphics[width=0.9\columnwidth]{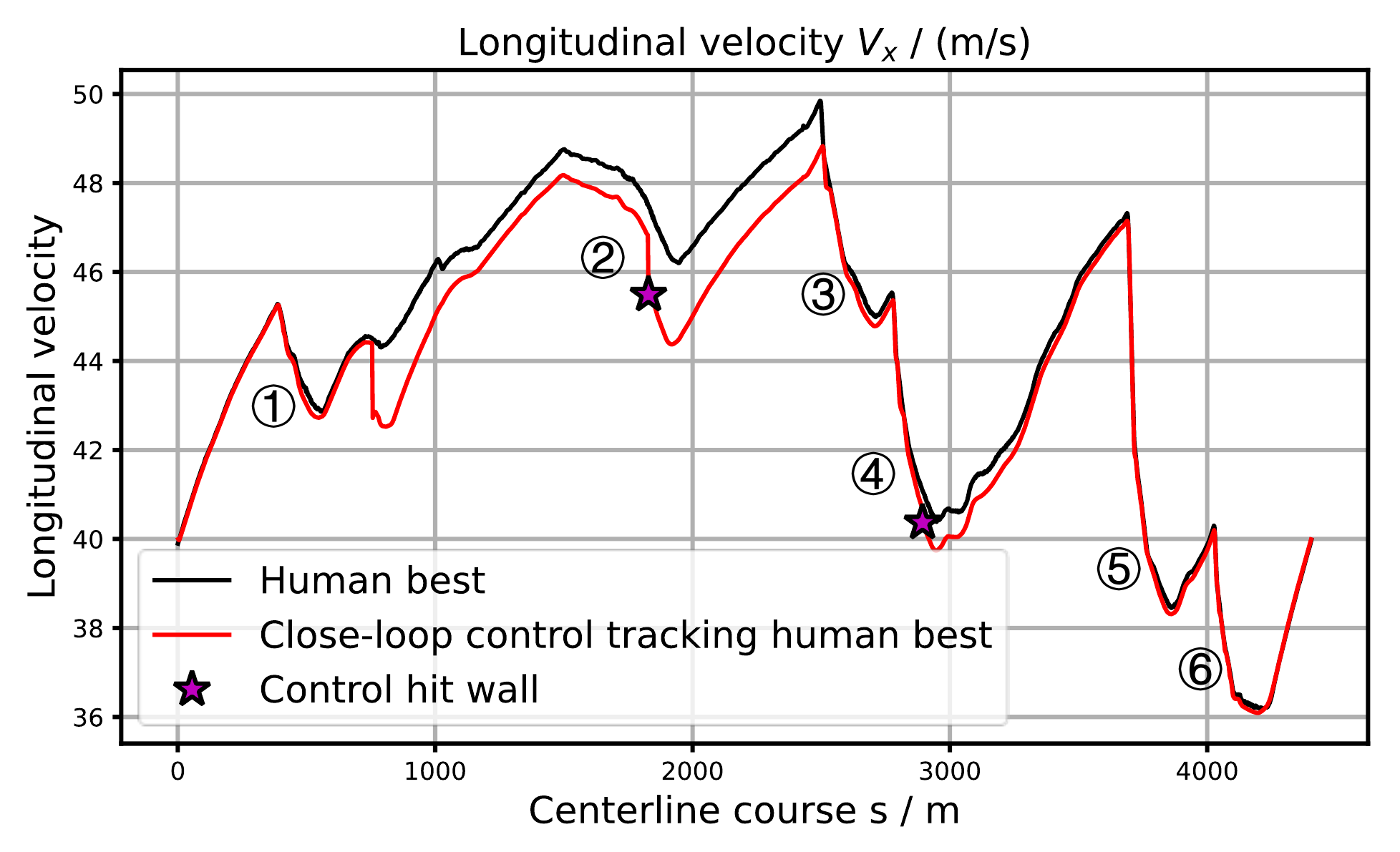}
    \caption{Tracking human best trajectory with coupled model MPC.}
    \label{fig:track_hum}
\end{figure}

Since we have access to the trajectory of the human-best recording, we conduct an experiment where we use the MPC controller to track the human-best trajectory to investigate our hypothesis. The results are summarized in Table~\ref{Tab:track_human_results} and Fig.~\ref{fig:track_hum}. The closed-loop lap time ($100.54s$) is slightly better than tracking the time-optimal trajectory ($100.83$). There is still a notable time gap ($1.11s$) between the reference lap time and the closed-loop lap time. One phenomenon we observed is that the race car could not track the longitudinal speed well. The longitudinal velocity error is larger than the error when tracking the time-optimal trajectory. In addition, the car crashed into the wall at the corner \textcircled{2} and \textcircled{4}. One possible reason is that the vehicle model is still not accurate enough due to some complicated unmodeled dynamics. Its effect was exaggerated when tracking the human-best trajectory, as the best human racer can push the car to its handling limit. 

In addition, we also observed some interesting improvements in comparison with tracking the time-optimal trajectory. First, the lateral tracking error is significantly smaller ($0.074s$ v.s. $0.158$). Also, the race car perfectly tracked the human-best trajectory in the corner \textcircled{1}, and in particular, the challenging S-shape corners \textcircled{5} and \textcircled{6} (Fig.~\ref{fig:track_hum}). \blue{ The longitudinal velocity error was much larger at these corners when tracking the time-optimal trajectory. It shows that the human-best trajectory achieves fewer tracking errors than the time-optimal trajectory in these regions. Therefore, we believe that the planner could also benefit from a more accurate vehicle dynamics model. The planned trajectory could be more easy-tracking and closer to the true optimal race line for the race car. }

A promising direction we would therefore like to explore in future research is to improve the modeling accuracy for \emph{both planning and control}. One challenge is that the remaining unmodeled dynamics could be difficult to identify and derive based on vehicle dynamics. Hence, we are interested in using data-driven approaches to learn a residual model compensating the modeling error of the current bicycle model from data. 

\section{Conclusion}
\label{Sec:Conclusion}

In this paper, we proposed a model-based control system for autonomous racing that achieved the top $0.95\%$ performance among human expert players in the high-fidelity racing platform Gran Turismo Sport. In particular, we pinpointed three crucial design challenges and proposed corresponding solutions, which all play important roles in advancing racing performance. First, we showed that it was necessary to consider the aerodynamic force and weight transfer effect in the vehicle model for both planning and control. Second, we proposed a two-stage time-optimal trajectory planning algorithm, where we use curvature-optimal planning to warmstart the non-convex time-optimal optimization problem. We showed that the proposed algorithm was able to find a time-optimal reference trajectory that was close to the human-best recording. Lastly, we showed that it was necessary for racing control to jointly optimize the pedal and steering commands by considering the coupled longitudinal and lateral vehicle dynamics in the MPC-based tracking controller.  

\section*{Acknowledgment}
This work was supported by Sony R$\&$D Center Tokyo and Sony AI. We would like to thank Kenta Kawamoto from Sony AI for their kind help and many fruitful discussions. We are also very grateful to Polyphony Digital Inc. for enabling this research and providing GT Sport framework. This material is based upon work supported by the National Science Foundation Graduate Research Fellowship Program under Grant No. DGE 1752814. Any opinions, findings, conclusions, or recommendations expressed in this material are those of the authors and do not necessarily reflect the views of the National Science Foundation.

\bibliographystyle{IEEEtran}
\bibliography{reference}

\appendices

\section{Simulation Environment} \label{Appendix: GTS}
In this section, we provide detailed information on the Gran Turismo Sport (GTS) environment, hardware setting, and the human player data collection process. GTS is a high-fidelity racing simulator with intricate vehicle models, complex tracks, and a talented, diverse human player base. Especially, realistic vehicle dynamics models are developed from a large amount of real-world racing data, real wind tunnel testing results, and human driver feedback\footnote{More detailed information is available at: \href{https://www.gtplanet.net/dr-kazunori-yamauchi-gives-lecture-gran-turismos-driving-physics-production/}{https://www.gtplanet.net/dr-kazunori-yamauchi-gives-lecture-gran-turismos-driving-physics-production/}.}. 
Leveraging this platform, we conduct experiments that compare the performance of model-based control to top human players \textit{in the same environment}, i.e., using the exact same vehicle and track. 

{\bf Scenarios.} We conduct experiments with the Mazda Demio XD Turing '15 on two distinct tracks. The first track is the Tokyo Expressway Central Outer Loop, which is notably bounded by walls around the track. During planning and control, it is important to ensure the planner considers a reasonable track width constraint to prevent collision.
The second track, Autopolis International Racing Course - Shortcut Course, lacks walls, allowing cars to drive off the course. However, driving off the track is undesirable as the friction outside the track area changes abruptly, altering vehicle dynamics. While human players often cut corners to reduce path length, we formulate the track width constraints to prevent this strategy in our model-based approach to avoid unpredictable shifts in the model. 

{\bf Human player recordings.} GTS holds daily ``time-trial'' attack competitions that are open publicly to all players. The top results with the lowest lap times of each day are recorded, including the full trajectories of the fastest laps. For example, the lap time distribution in Fig.~\ref{fig:laptime_distribution} was collected in a racing event ``GT Sport Daily Races'' in 2019\footnote{\href{https://www.gtplanet.net/gt-sport-daily-races-spa-day-feeling-bullish-and-tokyo-demio/}{https://www.gtplanet.net/gt-sport-daily-races-spa-day-feeling-bullish-and-tokyo-demio/}.}. Eventually, 28,000 successful trajectories in Race A setting (Mazda Demio on Tokyo Expressway) were collected. Our experiments are conducted on the same track using the same race car and under the same time-trial race setting; Thus, we ensure a fair comparison between the racing trials driven by human players and the developed autonomous racing software. 

{\bf Hardware.} Similar to~\cite{fuchs2021super}, we conduct experiments on a gaming computer Dell G7 with GeForce NVIDIA GTX 1080Ti GPUs that is connected to a PlayStation via an internet connection. The PlayStation and the accompanying inter-device communication impose additional costs, compared with other commonly used racing simulation environments. However, GTS's rich set of human player recordings brings unique benefits compared to other alternatives, which enables principled comparison between the performance of autonomous race cars and human racers. 

{\bf Control variables and states.} The control commands consist of a steering and a throttle/brake command. The vehicle states provided by the GTS API include some states that may be beyond the perception of human players (e.g., engine torque). To ensure a fair comparison, we exclusively utilize spatial states, namely,  2-dimensional position, orientation, and their derivatives. 
\color{black}

\section{Double-track Model} \label{Appendix: 4wheel model}

In autonomous driving, the bicycle model is widely adopted in model-based planning and control for simplicity. However, it neglects the vehicle's lateral movements, such as lateral weight transfer. In contrast, the double-track (i.e., four-wheel) model considers the vehicle dynamics and tire forces on \emph{four wheels}. The lateral dynamics of the double-track model are:
\begin{align}
    \dot{V}_{y}^{\text{4w}} =& \frac{1}{m}~\left\lbrack {\left( {F_{yFL} + F_{yFR}} \right){\cos(\delta)} + F_{yRL} + F_{yRR}} \right\rbrack - \dot{\psi}V_{x} \nonumber \\
    \ddot{\psi}^{\text{4w}} =& \frac{1}{I_{zz}}\left( {F_{yFL} + F_{yFR}} \right){\cos(\delta)}l_{f} - \left( {F_{yRL} + F_{yRR}} \right)l_{r} \nonumber \\
    &+ 0.5 \left( {F_{yFL} - F_{yFR}} \right){\sin(\delta)}l_{t}
\end{align}
where the superscript $\text{4w}$ denotes the variables from the four-wheel model; $l_t$ denotes the vehicle track width, i.e., the distance between the two front (rear) wheels. $F_{yFL}, F_{yFR}, F_{yRL}, F_{yRR}$ are lateral tire forces on the front/rear and left/right wheels, which are independently calculated by simplified Pacejka’s Magic Formula as Eq.~\eqref{Eqn:tire1} and \eqref{Eqn:tire2}. We assume that two front and two rear wheels have the same tire parameters and slip angles on two front wheels are identical. We also incorporate the lateral weight transfer by considering the change of wheel loads in Eq.~\eqref{Eqn: lateral weight transfer}. 

However, lateral weight transfer only has a marginal effect on the dynamics. Note that $ F_{yf} = F_{yFL} + F_{yFR}$ and $F_{yr} = F_{yRL} + F_{yRR}$. Therefore, the differences in lateral acceleration and yaw acceleration using the two models are 
\begin{align}
   & \dot{V}_y^{\text{4w}} - \dot{V}_y^{\text{bicycle}} = 0 \\
   & \ddot{\psi}^{\text{4w}} - \ddot{\psi}^{\text{bicycle}} = 
\frac{1}{I_{zz}}\left\lbrack {- 0.5l_{t} \cdot F_{y,\Delta W_{F}}} \right\rbrack{\sin(\delta)}
\end{align}
where, 
$F_{y,\Delta W_{F}} = \mu\Delta W_{F}D_{f}{\sin\left( {C_{f}{\arctan\left( {D_{f}\alpha_{f}} \right)}} \right)}$. It implies that the lateral weight transfer does not change the lateral acceleration, and only the sine component of front lateral tire force generated by $\Delta W_F$ influences the yaw acceleration, which is minor compared with the main yaw moment. Consequently, the bicycle model is a reasonable approximation of the double-track model because the lateral weight transfer has little impact on the lateral dynamics.

\color{black}

\section{Longitudinal tire acceleration model} \label{Appendix: long_force}
In GTS, the race cars are controlled with two control inputs, steering angle $\delta$ and combined pedal command $u_p$ (i.e., throttle and braking). The range of pedal is normalized to $[-1, 1]$. The vehicle brakes when $u_p \in [-1, 0)$ and accelerates when $u_p \in [0, 1]$. In the longitudinal dynamics model (Eq.~\eqref{Eqn:long_model}), the longitudinal velocity is controlled by the longitudinal tire acceleration $a_x$, which is actually determined by the pedal command through the powertrain and tire model. Therefore, the range of feasible $a_x$ depends on the vehicle state. To ensure feasible solutions in planning and control, we need to calculate the limits of $a_x$ according to the pedal command based on longitudinal tire forces on front and rear tires $F_{xf}$ and $F_{xr}$. Concretely, $a_x$ is related to the tire forces as:
\begin{equation} \label{equ:ax}
    a_x = \frac{F_{xf} + F_{xr}}{m}
\end{equation}
When the pedal command $u_p$ is non-zero, the vehicle engine or brake generates longitudinal tire forces through the powertrain dynamics. The powertrain dynamics in \cite{rajamani2011vehicle} is too sophisticated to be included in the vehicle model for the purpose of planning and control. To this end, we use linear regression to fit a linear engine model from data, which models the longitudinal slip ratios of the tires as functions of the vehicle states and pedal command:
\begin{equation}\label{equ:engine}
\renewcommand\arraystretch{1.5}
\begin{bmatrix}
\sigma_f \\ \sigma_r
\end{bmatrix}= 
\renewcommand\arraystretch{1.5}
\begin{bmatrix}
w_{\sigma f}^T \\ w_{\sigma r}^T
\end{bmatrix}
\renewcommand\arraystretch{1.5}
\begin{bmatrix}
\xi \\ u_p
\end{bmatrix}
\end{equation}
where $\sigma_f$ and $\sigma_r$ are the longitudinal slip ratios of the front and rear tires respectively. Then we model the longitudinal tire forces as a linear function of slip ratios as:
\begin{equation}\label{equ:long_tire}
    F_{xf}=C_{xf}\sigma_f, \quad F_{xr}=C_{xr}\sigma_r,
\end{equation}
where $C_{xf}$ and $C_{xr}$ are the longitudinal tire coefficients of the front and rear tires. Integrating Eq.~\eqref{equ:ax} to \eqref{equ:long_tire}, the  $a_x$ are calculated as :
\begin{equation}
    a_x = \textbf{M}
    \renewcommand\arraystretch{1.5}
    \begin{bmatrix}
    \xi \\ u_p
    \end{bmatrix}
\end{equation}
where $\textbf{M}= (C_{xf}w_{\sigma f}^T + C_{xr}w_{\sigma r}^T) / m$. Then the limits of $a_x$ are derive at the nominal states when $u_p = -1$ and $1$. In the controller, we also need to convert the solved $a_x$ back to the pedal command in the racing platform, so we calculate inverse of $\textbf{M}$  and the control pedal is as:
\begin{equation}
    u_p = \textbf{M}^{-1}(a_x - \textbf{M}\xi)
\end{equation}

\end{document}